\RequirePackage[svgnames,table]{xcolor}

\PassOptionsToPackage{skins,breakable}{tcolorbox}
\documentclass[11pt,letterpaper]{yalearxiv}

\usepackage{graphicx}
\usepackage{float,epstopdf}
\usepackage{bbm}

\usepackage{microtype}

\usepackage[numbers,sort&compress]{natbib}
\setcitestyle{square}

\usepackage{subcaption}
\usepackage{booktabs}

\usepackage{amsmath}
\usepackage{amssymb}
\usepackage{mathtools}
\usepackage{amsthm}
\usepackage{dsfont}
\usepackage{multicol}
\usepackage{makecell}
\usepackage{multirow} 
\usepackage{amsfonts} 
\usepackage{mathrsfs}
\usepackage[amssymb, thickqspace]{SIunits}
\usepackage{enumitem}
\usepackage{pgfplotstable}
\usepackage{lipsum}		

\usepackage{microtype}
\usepackage{graphicx}
\usepackage{booktabs} 
\usepackage[table]{xcolor}
\usepackage[normalem]{ulem} 

\usepackage{cases}
\usepackage{wrapfig}

\usepackage{url}

\definecolor{huskypurple}{HTML}{4B2E83}

\usepackage{titletoc}

\usepackage{listings}
\lstdefinestyle{promptstyle}{
  basicstyle=\ttfamily\footnotesize,
  breaklines=true,
  breakautoindent=false,
  breakindent=0pt,
  postbreak=\mbox{\textcolor{gray}{$\hookrightarrow$}\space},
  columns=fullflexible,
  keepspaces=true,
  frame=single,
  framesep=5pt,
  xleftmargin=6pt,
  xrightmargin=6pt,
  aboveskip=8pt,
  belowskip=8pt,
  showstringspaces=false,
}   

\AtEndPreamble{
    \usepackage[capitalize]{cleveref}
    \crefname{section}{Sec.}{Secs.}
    \Crefname{section}{Section}{Sections}
    \Crefname{table}{Table}{Tables}
    \crefname{table}{Tab.}{Tabs.}
}

\usepackage{amsthm}       
\usepackage{multirow}
\usepackage{colortbl}     
\usepackage{wrapfig}
\usepackage{algorithm}
\usepackage{algpseudocode}
\usepackage[skins,breakable]{tcolorbox}

\definecolor{mylinks}{RGB}{41, 171, 226} 
\definecolor{bestred}{RGB}{178, 24, 43}  

\newcommand{\sd}[1]{\,\raisebox{-0.25ex}{\tiny$\pm$#1}}
\newcommand{\scell}[2]{#1\sd{#2}}
\newcommand{\sbest}[2]{\cellcolor{besthl}\textcolor{bestred}{#1\sd{#2}}}
\newcommand{\sbestavg}[1]{\cellcolor{besthl}\textcolor{bestred}{#1}}

\definecolor{besthl}{RGB}{252, 228, 230}

\newtheoremstyle{thmplain}
  {0pt}{0pt}{\itshape}{}{\bfseries}{.}{.5em}{}
\newtheoremstyle{thmdefn}
  {0pt}{0pt}{}{}{\bfseries}{.}{.5em}{}

\theoremstyle{thmplain}
\newtheorem{theorem}{Theorem}
\newtheorem{proposition}{Proposition}

\newtheorem{corollary}{Corollary}
\theoremstyle{thmdefn}

\newtheorem{example}{Example}

\tcbset{
  thmstyle/.style={
    enhanced, breakable, boxrule=0.7pt, arc=2pt,
    left=8pt, right=8pt, top=6pt, bottom=6pt,
    before skip=10pt, after skip=10pt
  }
}
\tcolorboxenvironment{theorem}    {thmstyle, colframe=blue!55!black,   colback=blue!6}
\tcolorboxenvironment{proposition}{thmstyle, colframe=teal!70!black,   colback=teal!8}
\tcolorboxenvironment{lemma}      {thmstyle, colframe=violet!65!black, colback=violet!7}
\tcolorboxenvironment{corollary}  {thmstyle, colframe=orange!80!black, colback=orange!9}
\tcolorboxenvironment{definition} {thmstyle, colframe=green!45!black,  colback=green!8}
\tcolorboxenvironment{example}    {thmstyle, colframe=purple!60!black, colback=purple!6}
\tcolorboxenvironment{assumption} {thmstyle, colframe=red!55!black,    colback=red!5}

\AtEndPreamble{
  \crefname{theorem}{Thm.}{Thms.}         \Crefname{theorem}{Theorem}{Theorems}
  \crefname{proposition}{Prop.}{Props.}   \Crefname{proposition}{Proposition}{Propositions}
  \crefname{lemma}{Lem.}{Lems.}           \Crefname{lemma}{Lemma}{Lemmas}
  \crefname{corollary}{Cor.}{Cors.}       \Crefname{corollary}{Corollary}{Corollaries}
  \crefname{definition}{Def.}{Defs.}      \Crefname{definition}{Definition}{Definitions}
  \crefname{assumption}{Asm.}{Asms.}      \Crefname{assumption}{Assumption}{Assumptions}
  \crefname{example}{Ex.}{Exs.}           \Crefname{example}{Example}{Examples}
  \crefname{algorithm}{Alg.}{Algs.}       \Crefname{algorithm}{Algorithm}{Algorithms}
}

\definecolor{PrincetonOrange}{RGB}{245, 128, 37}

\usepackage{tabularx}
\newcolumntype{Y}{>{\raggedright\arraybackslash}X}

\makeatletter
\let\arxiv@orig@newl@bel\@newl@bel
\def\@newl@bel#1#2#3{%
  \arxiv@orig@newl@bel{#1}{#2}{#3}%
  \arxiv@orig@newl@bel{#1}{main-#2}{#3}}
\newcommand{\maketitlesupplementary}{%
  \clearpage
  \begin{center}
    \Large
    \textbf{\@title}\\
    \vspace{0.5em}Supplementary Material \\
    \vspace{1.0em}
  \end{center}}
\renewcommand{\tableofcontents}{%
  \section*{\contentsname}%
  \startcontents[supp]%
  \printcontents[supp]{}{1}{}}
\makeatother

\definecolor{cvprblue}{rgb}{0.21,0.49,0.74}
\hypersetup{colorlinks=true, allcolors=mylinks}
\usepackage[hyperpageref]{backref}

\title{Contingency Planning with Latent World Models under Irreversible Outcomes}

\runningtitle{Contingency Planning with Latent World Models under Irreversible Outcomes}

\author[1,*]{Abdul Monaf Chowdhury}
\author[1]{Syed Nazmus Sakib}
\author[2]{Ahanaf Hasan Ariq}
\author[1]{Chandak Chakma}
\author[1]{Nafiul Haque}
\author[3]{Shifat E. Arman}
\author[1]{Md Mehedi Hasan}
\affil[1]{\mbox{Department of Robotics and Mechatronics Engineering, University of Dhaka}}
\affil[2]{\mbox{Daffodil International School}}
\affil[3]{\mbox{Department of Engineering Science, University of Oxford}}

\correspondingauthor{Abdul Monaf Chowdhury, \href{mailto:monafabdul15@gmail.com}{\texttt{monafabdul15@gmail.com}}}

\begin{document}
\begin{abstract}

Latent world models plan in a compact latent space by predicting the outcomes of actions. However, these planners typically judge a plan by one deterministic rollout and its latent distance to the goal. When the environment is stochastic and some outcomes are irreversible, this fails. The rollout averages the possible outcomes and hides the risk, the distance ignores whether the goal is still reachable, and a fixed action sequence cannot react to new observations. To address this, we introduce \textbf{Contingent Latent Planning (CoLaP).} CoLaP keeps the world model frozen and trains from offline data without rewards. It samples several futures for each plan, scores them with a learned asymmetric time-to-goal that makes irreversible failures costly, and lets the plan branch on what the agent will observe. We evaluate CoLaP on nine stochastic navigation, reaching, and manipulation tasks, where it achieves the highest average success rate.



\end{abstract}

\maketitle
\suppressfloats[t]
\section{Introduction}
\label{sec:intro}

\begin{figure*}[t]
  \centering
  \includegraphics[width=\textwidth]{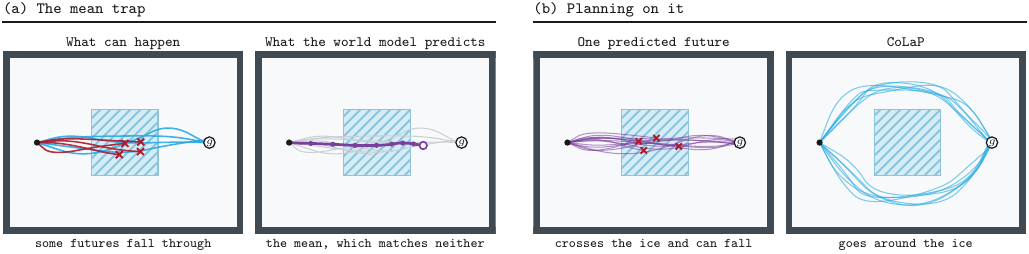}
  \caption{\textbf{The mean trap.} (a) A plan that crosses the ice has two outcomes. The agent either crosses, or it falls through and can no longer reach the goal $g$. Trained with squared error, the world model predicts the mean of these futures, which matches neither and shows no fall. (b) A planner that scores this one predicted future crosses the ice and falls in some episodes. CoLaP samples several futures, scores them by their expected time to the goal, and goes around the ice.}
  \label{fig:problem}
\end{figure*}

Latent world models predict the outcomes of actions in a compact latent space instead of in pixels~\cite{ha2018world,lecun2022path,hafner2019planet,hafner2025dreamerv3}. Many of them are trained offline on reward-free trajectories. A planner then searches for actions to reach the goal image~\cite{zhou2025dinowm,sobal2025pldm,assran2025vjepa2,maes2026lewm}. Recent models such as LeWorldModel (LeWM)~\cite{maes2026lewm} train end to end from pixels and plan far faster than world models built on foundation-model features.
The planner scores each candidate action sequence by the latent distance between its predicted outcome and the goal~\cite{zhou2025dinowm,sobal2025pldm,maes2026lewm}.


However, the planner assumes that each action sequence has a single outcome. In a stochastic environment, the same action sequence can lead to several outcomes, and some of them can be irreversible~\cite{moldovan2012safe,eysenbach2018leave,grinsztajn2021there}.
An agent crossing thin ice may fall through, and a pushed object may slip into a hole. Once such irreversible outcome happens, the goal becomes unreachable (\cref{fig:problem}). 
Since the planner sees only one predicted outcome per plan, its score hides this risk.


On such stochastic tasks, the planner fails in three ways. 
First, the world model is trained with squared error, so it predicts the mean of the possible outcomes~\cite{bishop2006pattern,mathieu2016deep}. On the ice example of \cref{fig:problem}, this mean lies between crossing and falling and matches neither~\cite{guo2022blade}. So, the planner never sees the fall. 
Second, the latent distance measures how close the agent is to the goal, not whether it can still reach it~\cite{moukheiber2026tempo,li2026rcaux}. So, a fallen agent near the goal can score better than a safe agent on a detour. 
Third, the planner searches for one fixed action sequence that must serve every outcome. 
So, the plan cannot change with what the agent will observe, and it sees no value in looking first and acting later~\cite{kaelbling1998planning}.



Our key insight is that all three failures have the same cause. The planner scores a plan as if the environment were deterministic. Instead, a plan should be scored by the expected time the agent needs to reach the goal, over all possible outcomes~\cite{bertsekas1991analysis}. This score weighs every outcome by how often it happens, so irreversible outcomes are no longer hidden. It also counts time instead of distance. As a result, an irreversible outcome costs the whole episode, even if the agent ends close to the goal. Finally, the agent may change its later actions after it observes an outcome. Looking first then pays off when it saves time later. We therefore keep the world model~\cite{maes2026lewm} frozen and build the planner around this score.


In this work, we introduce \textbf{Contingent Latent Planning (CoLaP)}, which estimates this score on top of a frozen world model~\cite{maes2026lewm}. It has three components, one for each failure. 
The first is a future sampler that draws several futures for each plan, so irreversible outcomes are no longer averaged away.
The second is an asymmetric time-to-goal that replaces the latent distance, so irreversible outcomes no longer look close to the goal.
The third is observation-conditioned branching that replaces the single action sequence, so the plan can change with what the agent will observe.  Our contributions are summarized as follows:



\begin{itemize}
  \item We show that latent planners fail under irreversible outcomes because they score each plan as if the environment were deterministic.
  \item We propose \textbf{CoLaP}, a planner that scores each plan by the expected time to reach the goal and learns which outcomes are irreversible from reward-free data alone.
  \item We evaluate CoLaP on nine stochastic navigation, reaching, and manipulation tasks with irreversible outcomes, where it achieves the highest average success rate.
\end{itemize}

\section{Related Work}
\label{sec:related}

\textbf{Latent world models.} Latent world models predict future states in a learned representation \cite{ha2018world,hafner2019planet,hafner2020dream,hafner2025dreamerv3,lecun2022path}. Some learn dynamics on frozen vision features \cite{zhou2025dinowm}, while others use reward-free offline trajectories \cite{sobal2025pldm,maes2026lewm}. LeWM scores a plan by comparing its predicted terminal embedding with the goal \cite{maes2026lewm}. A deterministic model predicts one future for each plan. Ensembles capture uncertainty through sampling \cite{chua2018pets} and generative models learn a distribution of outcomes \cite{daniel2026lpwm,denton2018stochastic,mathieu2016deep}. We keep the world model frozen and train a residual sampler on separate episodes. These episodes contain transitions the world model never saw during training. The sampler uses the energy score \cite{gneiting2007strictly} to learn the spread of outcomes around each prediction.

\textbf{Goal reaching with learned distances.} Standard embedding distances often fail to capture how hard it is to reach a goal \cite{kaelbling1993learning,pitis2020inductive}. To address this, quasimetric methods learn asymmetric distances that obey the triangle inequality \cite{liu2023mrn,wang2022learning,wang2023qrl}. QRL learns such a distance from trajectories under deterministic dynamics \cite{wang2023qrl}. Other methods learn temporal distances under stochastic dynamics \cite{myers2024learning} or replace a frozen planner’s cost with a symmetric temporal metric \cite{moukheiber2026tempo}. We use a learned quasimetric as cost-to-go and average it over sampled futures. Asymmetry lets the score distinguish entering a state from returning from it. This matters when an action leaves the agent with no easy route back. The average accounts for how often each outcome occurs.

\textbf{Contingency planning.} Contingency planners share the initial actions and choose different continuations after the outcomes become clear \cite{hardy2013contingency,rhinehart2021contingencies}. Prior methods branch over predicted scenes \cite{cui2021lookout} or build scenario specific plans with shared initial controls \cite{chen2022branchmpc}. We apply this structure to a frozen latent model. Our sampler supplies the possible futures and we group them by their predicted observations at the branch point. Futures in the same group share a continuation. This prevents the planner from choosing different actions for outcomes the agent cannot yet distinguish. We execute only the first two chunks of the root plan and then replan. The alternative helps score whether an observation would let the agent choose a better continuation.

\section{Preliminaries}
\label{sec:prelims}

\paragraph{Problem setting.}
We consider goal-conditioned control from pixels in a fully offline and reward-free setting
~\cite{kaelbling1993learning,levine2020offline,park2025ogbench,zhou2025dinowm,sobal2025pldm,maes2026lewm}. We are given a dataset $\mathcal{D}$ of 
trajectories made of image observations $\boldsymbol{o}_{1:T}$ and actions $\boldsymbol{a}_{1:T}$, with no rewards 
and no further access to the environment.
As in LeWM~\cite{maes2026lewm}, each action $\boldsymbol{a}_t$ is an \emph{action chunk}~\cite{chi2023diffusionpolicy,li2025qchunking} 
of five consecutive actions. So, one time step covers five environment actions.
At test time, the agent starts from an observation $\boldsymbol{o}_1$ and receives a goal image
$\boldsymbol{o}_g$. It succeeds if it reaches the goal before the episode ends.


\paragraph{Latent world model.}
A latent world model~\cite{ha2018world,lecun2022path,hafner2019planet,hafner2020dream,hansen2022tdmpc,hafner2025dreamerv3} predicts the future 
in a learned embedding space instead of in pixels~\cite{assran2023ijepa,bardes2024vjepa,assran2025vjepa2}. It has an encoder and a predictor,
\begin{equation}
  \boldsymbol{z}_t = \mathrm{enc}_{\theta}(\boldsymbol{o}_t), \qquad \hat{\boldsymbol{z}}_{t+1} = \mathrm{pred}_{\phi}(\boldsymbol{z}_t, \boldsymbol{a}_t),
  \label{eq:lewm}
\end{equation}
where $\boldsymbol{z}_t \in \mathbb{R}^{D}$ is the embedding of the observation $\boldsymbol{o}_t$ and 
$\hat{\boldsymbol{z}}_{t+1}$ is the predicted embedding of the next observation $\boldsymbol{o}_{t+1}$.
We use LeWorldModel (LeWM)~\cite{maes2026lewm}, which trains both networks 
end-to-end from pixels with
\begin{equation}
  \mathcal{L}_{\text{LeWM}} \triangleq
  \lVert \hat{\boldsymbol{z}}_{t+1} - \boldsymbol{z}_{t+1} \rVert_2^2 + \lambda\, \mathrm{SIGReg}(\boldsymbol{Z}).
  \label{eq:lewm-loss}
\end{equation}
The first term pulls each prediction toward the true next embedding. 
The second term, SIGReg~\cite{balestriero2025lejepa}, prevents representation collapse~\cite{chen2020simclr,grill2020byol,zbontar2021barlow,bardes2022vicreg,jing2022understanding} 
by pushing the embeddings $\boldsymbol{Z}$ toward an isotropic Gaussian.
So, the encoder cannot map every observation to the same embedding.
The predictor also reads the previous embeddings, which we leave implicit as LeWM does. 
Given these embeddings and an action, it returns a single next embedding.
We train one LeWM per task~\cite{maes2026swm} and keep it frozen.
Its architecture and training details are in the appendix.


\begin{figure*}[t]
  \centering
  \includegraphics[width=\textwidth]{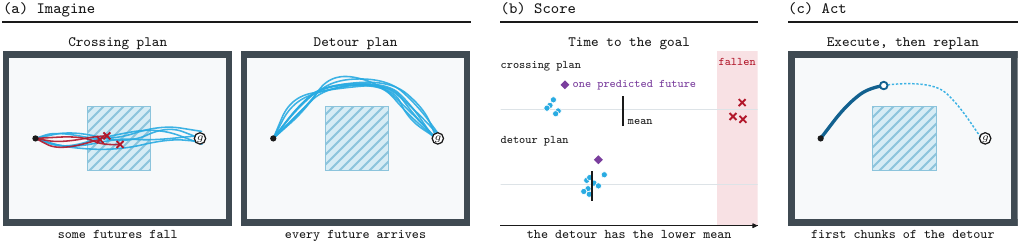}
  \caption{\textbf{One planning step of CoLaP.} (a) The sampler imagines several futures of each candidate plan, and some futures of the crossing fall through the ice. (b) The time-to-goal scores each future, and a fall costs the rest of the episode. The average over the futures ranks the detour first, while one predicted future ranks the crossing first. (c) The agent executes the first chunks of the best plan and then replans. Branching is shown in \cref{fig:components}c.}
  \label{fig:step}
\end{figure*}

\paragraph{Latent planning.}
LeWM~\cite{maes2026lewm} plans in its embedding space. 
The predictor rolls out a candidate plan of $H$ action chunks 
$\boldsymbol{a}_{1:H}$, one chunk at a time. It starts from $\hat{\boldsymbol{z}}_1 = \mathrm{enc}_{\theta}(\boldsymbol{o}_1)$ 
and applies \cref{eq:lewm} to its own predictions.
The planner compares the last prediction with the goal embedding 
$\boldsymbol{z}_g = \mathrm{enc}_{\theta}(\boldsymbol{o}_g)$ through the cost $\mathcal{C}(\hat{\boldsymbol{z}}_{H+1}) = \lVert \hat{\boldsymbol{z}}_{H+1} - \boldsymbol{z}_g \rVert_2^2$, and solves
\begin{equation}
  \boldsymbol{a}^{*}_{1:H} = \operatorname*{arg\,min}_{\boldsymbol{a}_{1:H}} \, \mathcal{C}(\hat{\boldsymbol{z}}_{H+1}).
  \label{eq:lewm-plan}
\end{equation}
The planner solves \cref{eq:lewm-plan} with the Cross-Entropy Method 
(CEM)~\cite{rubinstein1999cem,deboer2005tutorial}. 
CEM iteratively samples plans from a Gaussian, scores them with $\mathcal{C}$, 
and refits the Gaussian to the lowest-cost plans~\cite{chua2018pets,hafner2019planet,zhou2025dinowm}.
The agent executes the first few chunks of $\boldsymbol{a}^{*}_{1:H}$,
observes the result, and replans from the new observation, 
as in model predictive control (MPC)~\cite{richalet1978model,garcia1989mpc}.
So, each plan is scored by a single predicted future and its distance to the goal.






\section{Contingent Latent Planning}
\label{sec:method}

In this section, we introduce our method Contingent Latent Planning (CoLaP). We first
define the problem and show where latent planning fails on it. We then describe
three components that address these failures: a cross-fitted future sampler
(\cref{sec:sampler}), a learned expected time-to-goal (\cref{sec:ttg}), and
observation-conditioned branching (\cref{sec:contingency}). All three are added to a frozen LeWM \cite{maes2026lewm}.


\subsection{The Problem}
\label{sec:problem}

The planner of \cref{sec:prelims} scores a plan by one predicted future and its distance to the goal. But what happens when the environment is stochastic? 
The same action chunk can then lead to different outcomes, and some of them are irreversible~\cite{moldovan2012safe,eysenbach2018leave,grinsztajn2021there}. Once such an outcome happens, the goal can no longer be reached.

\begin{figure*}[t]
  \centering
  \includegraphics[width=\textwidth]{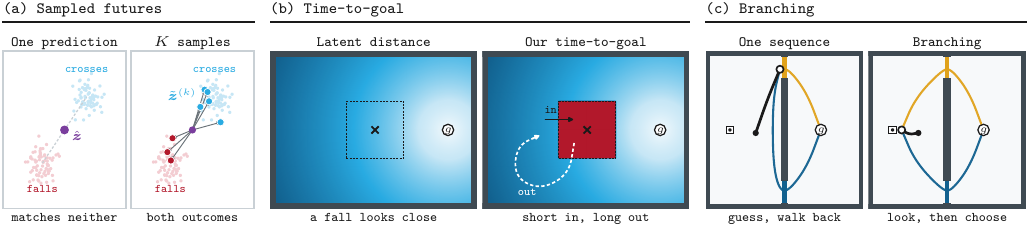}
  \caption{\textbf{The three components of CoLaP.} (a) LeWM predicts one embedding $\hat{\boldsymbol{z}}$, the mean of the two outcomes, and the sampler turns it into $K$ samples $\tilde{\boldsymbol{z}}^{(k)}$ that cover both. (b) The cost of reaching $g$ from each point, with a fallen agent on the ice. The latent distance places the fall close to $g$, while our time-to-goal makes the way in short and the way out long. (c) Each color is one lock state, and its path takes the door of its color. One action sequence must guess a door, while branching checks the switch first.}
  \label{fig:components}
\end{figure*}

We judge a policy $\pi$ by its expected time to the goal~\cite{bertsekas1991analysis}. Let $\tau \le L$ be the
number of chunks the agent takes to reach the goal, where $L$ is the episode
length and $\tau = L$ if the agent never arrives. The agent should minimize the expected time to goal,
\begin{equation}
  \min_{\pi} \; \mathbb{E}_{\pi}\!\left[\tau\right], \qquad
  \boldsymbol{a}_t \sim \pi(\cdot \mid \boldsymbol{o}_{1:t}, \boldsymbol{o}_g).
  \label{eq:objective}
\end{equation}
where the expectation is over the stochastic transitions of the environment. An irreversible failure sets $\tau = L$, so it costs the whole episode. Consider the following examples,



\begin{example}[Ice]
\label{ex:ice}
The direct path to the goal takes 10 chunks and crosses a patch of ice. Halfway, the agent falls through with probability $0.5$ and is stuck for the rest of the episode. So, $\tau = L = 50$ in that case. 
The direct path then takes $0.5 \cdot 10 + 0.5 \cdot 50 = 30$ chunks in expectation, while a detour around the ice takes 20. So, the detour is better.
\end{example}

\begin{example}[Locked door]
\label{ex:door}
Two doors lead to the goal, each in 10 chunks, and one of them is locked at random. If the agent picks the locked door, it must walk to the other one, which costs 8 more chunks. Guessing a door then takes: $0.5 \cdot 10 + 0.5 \cdot (10+8) = 14$ chunks in expectation. A switch that shows which door is locked costs a detour of 2 chunks. After which the agent walks straight to the open door in $2 + 10 = 12$. So, checking the switch is better.
\end{example}


The planner of \cref{sec:prelims} takes the ice and skips the switch, for three
reasons.

First, it sees one future. LeWM \cite{maes2026lewm} is trained with squared error, so it predicts the mean of the next embedding~\cite{bishop2006pattern}. 
On the ice, this mean lies between crossing and falling and matches neither~\cite{mathieu2016deep,denton2018stochastic,guo2022blade}. So, the planner never sees the fall. \Cref{sec:sampler} replaces it with sampled futures.

Second, it scores distance, not arrival~\cite{moukheiber2026tempo,li2026rcaux}. A fallen agent 5 chunks from the goal can score better than an agent 5 chunks into the detour, yet it never arrives.
\Cref{sec:ttg} replaces the distance with a learned expected time-to-goal.

Third, it plans one action sequence for every future. That sequence cannot depend on which door the switch shows. So, the planner sees the visit to the switch as 2 wasted chunks and never checks. \Cref{sec:contingency} lets the plan branch on what the agent will observe.




\subsection{Cross-fitted Future Sampler}
\label{sec:sampler}

In \cref{ex:ice}, an agent on the ice either falls through or crosses safely. To see the risk, the planner needs futures that keep the outcomes apart. We add a small sampler $g_{\psi}$ on top of the frozen predictor \cite{maes2026lewm} that turns each prediction into a sample of the next embedding~\cite{hafner2019planet,daniel2026lpwm}. We then feed the sample back into the predictor.

For a plan $\boldsymbol{a}_{1:H}$ and noise $\boldsymbol{\epsilon}^{(k)}_t \sim \mathcal{N}(\mathbf{0}, \boldsymbol{I})$, the $k$-th future $\tilde{\boldsymbol{z}}^{(k)}_{2:H+1}$ is generated for $t = 1, \dots, H$ by
\begin{equation}
  \begin{aligned}
    \hat{\boldsymbol{z}}^{(k)}_{t+1} &= \mathrm{pred}_{\phi}\big(\tilde{\boldsymbol{z}}^{(k)}_t, \boldsymbol{a}_t\big),\\
    \tilde{\boldsymbol{z}}^{(k)}_{t+1} &= \hat{\boldsymbol{z}}^{(k)}_{t+1} + g_{\psi}\big(\hat{\boldsymbol{z}}^{(k)}_{t+1},
    \boldsymbol{\epsilon}^{(k)}_t\big),
  \end{aligned}
  \label{eq:sampler}
\end{equation}
starting from the embedding of the current frame, $\tilde{\boldsymbol{z}}^{(k)}_1 = \mathrm{enc}_{\theta}(\boldsymbol{o}_1)$. 
Since each sample is fed back, every later prediction in the future builds on it. 
Repeating \cref{eq:sampler} with $K$ independent noise sequences gives $K$ futures for the same plan (\cref{fig:components}a).

We train $g_{\psi}$ with the energy score (ES)~\cite{gneiting2007strictly,szekely2013energy},
\begin{equation}
  \mathcal{L}_{\text{ES}} \triangleq
  \mathbb{E}\big[\lVert \tilde{\boldsymbol{z}}_{t+1} - \boldsymbol{z}_{t+1} \rVert_2\big]
  - \tfrac{1}{2}\, \mathbb{E}\big[\lVert \tilde{\boldsymbol{z}}_{t+1} - \tilde{\boldsymbol{z}}'_{t+1} \rVert_2\big],
  \label{eq:es}
\end{equation}
where $\tilde{\boldsymbol{z}}_{t+1}$ and $\tilde{\boldsymbol{z}}'_{t+1}$ are two samples with independent noise for the same prediction $\hat{\boldsymbol{z}}_{t+1}$. Here, $\hat{\boldsymbol{z}}_{t+1}$ is predicted from real frames, and $\boldsymbol{z}_{t+1}$ is the embedding of the real next frame. The first term of \cref{eq:es} pulls the samples toward the real outcome, and the second pushes them apart. Given the prediction $\hat{\boldsymbol{z}}_{t+1}$, the expected score is lowest only when the samples follow the true distribution of the embedding $\boldsymbol{z}_{t+1}$. 
So, the sampler $g_{\psi}$ learns the distribution of LeWM's error $\boldsymbol{z}_{t+1} - \hat{\boldsymbol{z}}_{t+1}$, and the samples spread over the outcomes instead of sitting at the prediction.


However, LeWM's error is smaller on its own training data than on new episodes. It sees each training sequence for many epochs and can memorize its outcome~\cite{zhang2017understanding,arpit2017closer}.
So, a sampler trained on these sequences would underestimate the spread of the outcomes. To prevent this, we cross-fit~\cite{chernozhukov2018double,lei2018distribution}. 
We train LeWM and $g_{\psi}$ on disjoint sets of episodes from the same behavior policy, where $g_{\psi}$ uses 5{,}000 episodes that LeWM never sees.

In practice, sequences in which a random outcome can occur next, such as an agent at the edge of the ice, are rare. We oversample them into half of the training sequences in every epoch. We find them from the simulator's logged position during training. Since the oversampling never looks at the outcome, the outcome distribution of each sequence stays unbiased~\cite{shimodaira2000improving}. At planning time, we draw $K = 8$ futures per plan, and all candidates in a CEM~\cite{rubinstein1999cem,deboer2005tutorial} iteration share the same noise~\cite{ng2000pegasus,glasserman1992some}. So, any difference between candidates comes from their actions and not from sampling luck. 

\subsection{Learned Expected Time-to-Goal}
\label{sec:ttg}

Going back to \cref{ex:ice}, a fallen agent can lie close to the goal, yet it never arrives. So, the planner should score how soon a future reaches the goal, not how close it ends. We learn a time-to-goal $d(\boldsymbol{z}, \boldsymbol{z}')$ that estimates the fewest action chunks needed to go from the embedding $\boldsymbol{z}$ to the embedding $\boldsymbol{z}'$~\cite{kaelbling1993learning,hartikainen2020dynamical,eysenbach2019search,myers2024learning}. For the $k$-th future of a plan $\boldsymbol{a}_{1:H}$, we define
\begin{equation}
  \begin{aligned}
    J^{(k)}(\boldsymbol{a}_{1:H}) &= \sum_{t=1}^{H}
      \min\big(d(\tilde{\boldsymbol{z}}^{(k)}_{t+1}, \boldsymbol{z}_g),\, 1\big)\\
      &\quad + d\big(\tilde{\boldsymbol{z}}^{(k)}_{H+1}, \boldsymbol{z}_g\big),\\
    V(\boldsymbol{a}_{1:H}) &= \frac{1}{K} \sum_{k=1}^{K} J^{(k)}(\boldsymbol{a}_{1:H}).
  \end{aligned}
  \label{eq:ttg}
\end{equation}
Here, $J^{(k)}$ is the time to the goal along the $k$-th future. For each action chunk of the plan, the first term counts the chunks spent before the future reaches the goal. The second term adds the chunks the future still needs after the plan ends, which is about $0$ if it has already arrived. 
The average of $J^{(k)}$ over the $K$ futures gives $V$, which estimates the expected time in \cref{eq:objective}. Planner minimizes $V$ instead of the cost $\mathcal{C}$ in \cref{eq:lewm-plan}.




For $J^{(k)}$ to penalize a fall, $d$ must place a fallen agent far from the goal. From \cref{ex:ice}, let the agent on the ice be at the embedding $\boldsymbol{z}$ and fall through in one action chunk to the embedding $\boldsymbol{z}'$. Falling takes one chunk, so $d(\boldsymbol{z}, \boldsymbol{z}') \le 1$. Getting back out is impossible, so the true time from $\boldsymbol{z}'$ back to $\boldsymbol{z}$ is unbounded. A symmetric time-to-goal $d$ cannot express this, since it sets $d(\boldsymbol{z}', \boldsymbol{z}) = d(\boldsymbol{z}, \boldsymbol{z}') \le 1$. The time from $\boldsymbol{z}'$ to the goal is at most the time from $\boldsymbol{z}'$ to $\boldsymbol{z}$ plus the time from $\boldsymbol{z}$ to the goal. With a symmetric $d$, this gives
\begin{equation}
  d(\boldsymbol{z}', \boldsymbol{z}_g) \le d(\boldsymbol{z}', \boldsymbol{z}) + d(\boldsymbol{z}, \boldsymbol{z}_g) \le 1 + d(\boldsymbol{z}, \boldsymbol{z}_g).
  \label{eq:bound}
\end{equation}

So, with a symmetric $d$, falling through the ice costs at most one extra action chunk, although the fallen agent never reaches the goal. We therefore need an asymmetric time-to-goal $d$, with a small time into the fall and a large time out of it (\cref{fig:components}b).


We model time-to-goal $d$ as a quasimetric \footnote{\textbf{Quasimetric:} A distance that obeys the triangle inequality but does not need to be symmetric~\cite{wang2022learning,pitis2020inductive}.} with a metric residual network (MRN)~\cite{liu2023mrn},
\begin{equation}
  d(\boldsymbol{z}, \boldsymbol{z}') = \lVert f(\boldsymbol{z}) - f(\boldsymbol{z}') \rVert_2
  + \max_i \big[h_i(\boldsymbol{z}) - h_i(\boldsymbol{z}')\big]_+
  \label{eq:mrn}
\end{equation}
where $f$ and $h$ are two output heads of one network and $[x]_+ = \max(x, 0)$. The first term is symmetric and the second is not. Both obey the triangle inequality, so time-to-goal $d$ is a quasimetric for any network weights. We train $d$ with quasimetric RL (QRL)~\cite{wang2023qrl},
\begin{equation}
  \begin{aligned}
    \max_{d}\;\; & \mathbb{E}_{\boldsymbol{z}, \boldsymbol{z}'}\big[\varphi\big(d(\boldsymbol{z}, \boldsymbol{z}')\big)\big]\\
    \text{s.t.}\;\; & \mathbb{E}_{\boldsymbol{z}_t, \boldsymbol{z}_{t+1}}
      \big[\,[d(\boldsymbol{z}_t, \boldsymbol{z}_{t+1}) - 1]_+^2\,\big] \le \eta^2
  \end{aligned}
  \label{eq:qrl}
\end{equation}
where $\boldsymbol{z}$ and $\boldsymbol{z}'$ embed two random frames, and $\boldsymbol{z}_t$ and $\boldsymbol{z}_{t+1}$ embed two consecutive frames of the same episode. The objective pushes $d$ up between random frames, while the constraint keeps $d$ at most $1$ between consecutive frames, up to a small tolerance $\eta$.
Since $d$ obeys the triangle inequality, it then learns the fewest chunks needed to go from one frame to another in the data.
Here, $\varphi$ increases and flattens above a cap $d_{\max}$, so pairs that are already far apart are barely pushed further. We solve \cref{eq:qrl} by dual ascent~\cite{boyd2011distributed} on a Lagrange multiplier, as QRL does.


In the data, no episode recovers from an irreversible outcome such as a fall. 
So, the constraint keeps the step into it short, $d(\boldsymbol{z}, \boldsymbol{z}') \le 1$, but no episode ever leads back from $\boldsymbol{z}'$.
Nothing then stops the objective from pushing $d(\boldsymbol{z}', \boldsymbol{z})$ up. In this way, $d$ learns which outcomes are irreversible from the data alone, without failure labels~\cite{grinsztajn2021there}. The mean in \cref{eq:ttg} then weighs each irreversible outcome by how often it happens, so CoLaP needs no extra risk parameter~\cite{rockafellar2000optimization,chow2015risk}.




The planner evaluates $d$ on sampled embeddings $\tilde{\boldsymbol{z}}$, which lie near real embeddings but not exactly on them. Trained only on real embeddings, $d$ can read such a small offset as several chunks. To prevent this, we add a second constraint to \cref{eq:qrl} with its own multiplier,
\begin{equation}
  \mathbb{E}\big[\,[d(\boldsymbol{z} + \boldsymbol{\xi}, \boldsymbol{z}) - \delta]_+^2
  + [d(\boldsymbol{z}, \boldsymbol{z} + \boldsymbol{\xi}) - \delta]_+^2\,\big] \le \eta^2
  \label{eq:noise}
\end{equation}
where $\boldsymbol{z}$ embeds a random frame and $\boldsymbol{\xi} \sim \mathcal{N}(\mathbf{0}, \sigma^2 \boldsymbol{I})$ is small Gaussian noise~\cite{bishop1995training}, with $\sigma$ drawn for each pair. When both terms are at most $\delta$, the triangle inequality gives $|d(\boldsymbol{z} + \boldsymbol{\xi}, \boldsymbol{z}_g) - d(\boldsymbol{z}, \boldsymbol{z}_g)| \le \delta$ for any goal $\boldsymbol{z}_g$. 
So, a small offset barely changes the time to the goal by at most $\delta$ chunks.

In practice, we train $d$ on the frames of LeWM's training episodes. Since $d$ never sees LeWM's predictions, it does not need cross-fitting.




\subsection{Observation-Conditioned Branching}
\label{sec:contingency}

For \cref{ex:door}, a good plan first checks the switch and then walks to the open door. Its later action depend on what the agent observes on the way, which a single action sequence cannot express. So, we let each candidate plan branch on what the agent will observe~\cite{hardy2013contingency,cui2021lookout,rhinehart2021contingencies,chen2022branchmpc,bouzidi2024motion}.

Each candidate plan is now a pair, a root plan $\boldsymbol{a}_{1:H}$ and an alternative plan $\boldsymbol{a}'_{1:H}$. Every future follows the root for the first $b$ chunks. The agent then observes $\boldsymbol{o}_{b+1}$ and either keeps the root or switches to the alternative. Futures that are close in embedding space look the same to the agent at this point, so they must continue in the same way~\cite{rockafellar1991scenarios,birge2011introduction}.
We therefore split the $K$ futures into two groups, $\mathcal{G}_1$ and $\mathcal{G}_2$, by 2-means~\cite{lloyd1982least} on their embeddings $\tilde{\boldsymbol{z}}^{(k)}_{b+1}$. 
Each group either keeps the root or switches, whichever gives it the lower total time-to-goal,
\begin{equation}
  \begin{aligned}
    V_b &= \frac{1}{K} \sum_{j=1}^{2} \min\Big\{
      \sum_{k \in \mathcal{G}_j} J^{(k)}(\boldsymbol{a}_{1:H}),\\
      &\qquad \sum_{k \in \mathcal{G}_j} J^{(k)}(\boldsymbol{a}_{1:b}, \boldsymbol{a}'_{b+1:H})\Big\}
  \end{aligned}
  \label{eq:branch}
\end{equation}
where $(\boldsymbol{a}_{1:b}, \boldsymbol{a}'_{b+1:H})$ follows the root for the first $b$ chunks and the alternative after that. Its futures continue from the root's samples at chunk $b$, with the same noise. We branch after chunk $b \in \{1, 2\}$ and take $V = \min_b V_b$ (\cref{fig:components}c).


CEM~\cite{rubinstein1999cem,deboer2005tutorial} searches the root and the alternative together, as one plan of $2H$ chunks. Since each group can keep the root, $V_b$ never exceeds the value $V(\boldsymbol{a}_{1:H})$ of the root alone. The agent executes the first two chunks of the root and then replans from the new observation. So, the alternative is never executed directly. Instead, it lets the planner value a root that gathers information~\cite{howard1966information,kaelbling1998planning}, such as the visit to the switch of \cref{ex:door}. The architecture and all settings are in the appendix.



\subsection{Overall Objective}
\label{sec:objective}

CoLaP replaces the planning problem of \cref{eq:lewm-plan} with
\begin{equation}
  (\boldsymbol{a}^{*}_{1:H}, \boldsymbol{a}'^{*}_{1:H}) = \operatorname*{arg\,min}_{\boldsymbol{a}_{1:H},\, \boldsymbol{a}'_{1:H}}
  \; \min_{b \in \{1, 2\}} V_b(\boldsymbol{a}_{1:H}, \boldsymbol{a}'_{1:H}).
  \label{eq:full}
\end{equation}
CEM scores each candidate pair in three steps. The sampler rolls out $K$ futures (\cref{eq:sampler}), the time-to-goal $d$ scores them (\cref{eq:ttg}), and each group of futures keeps the root or switches (\cref{eq:branch}) based on what the agent will observe. CEM solves \cref{eq:full}, and the agent executes the first two chunks of $\boldsymbol{a}^{*}_{1:H}$ before it replans, as in \cref{sec:prelims}. 
LeWM stays frozen, and the sampler $g_{\psi}$ and the time-to-goal $d$ are trained once per task, before planning. The algorithm for CoLaP is presented in the appendix.

\section{Experiments}
\label{sec:exp}


We evaluate CoLaP on nine stochastic tasks built on the environments of LeWM~\cite{maes2026lewm,maes2026swm}. 
\Cref{sec:exp-main} compares CoLaP with nine baselines on all of them. \Cref{sec:exp-components} then removes each component of CoLaP to see what it contributes, and \cref{sec:exp-design} tests three design choices.

\begin{figure}[t]
  \centering
  \includegraphics[width=\linewidth]{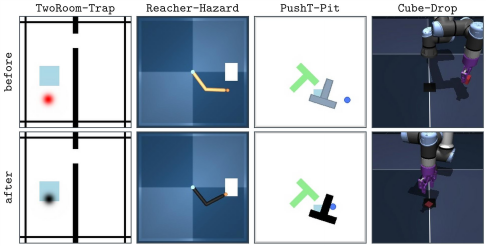}
  \caption{\textbf{Tasks.} Observations from four of the nine tasks, as LeWM sees them. Each column shows one training episode before (top) and after (bottom) its irreversible outcome. The agent falls through the ice, the arm breaks on the patch, the block falls into the pit, and the cube slips into the hole. After the outcome, the goal can no longer be reached.}
  \label{fig:tasks}
\end{figure}

\paragraph{Baselines.}
Our baselines come from three families, listed below:
\begin{itemize}[leftmargin=1.4em, itemsep=1pt, topsep=3pt, parsep=0pt]
  \item \textbf{LeWM planners}: LeWM's published planner~\cite{maes2026lewm}, one predicted future, MC-dropout~\cite{gal2016dropout} and TEMPO~\cite{moukheiber2026tempo}
  \item \textbf{Our futures in open loop}: the sampler in open loop, and CVaR~\cite{rockafellar2000optimization}
  \item \textbf{Offline Goal-conditioned RL}: GCBC~\cite{ghosh2021gcsl}, GCIVL~\cite{park2023hiql} and GCIQL~\cite{kostrikov2022iql}
\end{itemize}
GCBC, GCIVL and GCIQL are the goal-conditioned reinforcement learning variants of BC, IVL and IQL~\cite{park2025ogbench}. We train them on the frozen LeWM embeddings. All planners use the same frozen LeWM and plan one action sequence. We report our results with standard deviations, and the best results are marked in \textcolor{bestred}{\textbf{RED}}. Details of the implementation, the baselines and the hyperparameters are in the appendix.

\begin{table*}[t]
\centering
\caption{\textbf{Evaluation on nine stochastic tasks.} We report success rate (\%) and standard deviation over four seeds, with 50 episodes per seed. All of the baselines use 8 futures, as in CoLaP. Best results are marked in \textcolor{bestred}{\textbf{RED}} and highlighted.}
\label{tab:main}
\footnotesize\ttfamily
\setlength{\tabcolsep}{3pt}
\resizebox{\textwidth}{!}{%
\begin{tabular}{llcccccccccc}
\toprule
 & & \multicolumn{4}{c}{LeWM planners} & \multicolumn{2}{c}{Our futures, open loop} & \multicolumn{3}{c}{Offline GCRL} & \\
\cmidrule(lr){3-6}\cmidrule(lr){7-8}\cmidrule(lr){9-11}
Environment & Task & LeWM & One future & MC-dropout & TEMPO & Sampler & CVaR & GCBC & GCIVL & GCIQL & \textcolor{bestred}{\textbf{CoLaP}} \\
\midrule
\multirow{4}{*}{Navigation}
 & TwoRoom-Trap   & \scell{75.0}{6.0} & \scell{77.5}{5.7} & \scell{96.5}{1.9} & \scell{78.0}{4.3} & \scell{98.5}{1.0} & \scell{99.5}{1.0} & \scell{40.5}{4.4} & \scell{67.0}{6.0} & \scell{42.5}{8.7} & \sbest{99.5}{1.0} \\
 & TwoRoom-Bridge & \scell{39.0}{8.2} & \scell{57.5}{2.5} & \scell{59.5}{4.4} & \scell{45.5}{6.8} & \scell{76.0}{7.1} & \scell{79.0}{4.8} & \scell{42.5}{5.0} & \scell{73.5}{3.4} & \scell{29.5}{6.6} & \sbest{83.5}{1.9} \\
 & TwoRoom-Slip   & \scell{93.5}{3.0} & \scell{100.0}{0.0} & \scell{100.0}{0.0} & \scell{94.0}{2.8} & \scell{100.0}{0.0} & \scell{99.5}{1.0} & \scell{46.5}{10.6} & \scell{100.0}{0.0} & \scell{100.0}{0.0} & \sbest{100.0}{0.0} \\
 & TwoRoom-LockedDoor & \scell{0.0}{0.0} & \scell{60.5}{7.9} & \scell{68.5}{7.7} & \scell{35.5}{4.4} & \scell{59.0}{6.6} & \scell{57.5}{4.4} & \scell{6.0}{3.3} & \scell{0.0}{0.0} & \scell{7.0}{7.6} & \sbest{71.0}{6.1} \\
\midrule
\multirow{2}{*}{Reaching}
 & Reacher-Thin   & \scell{60.5}{2.5} & \scell{67.5}{2.5} & \scell{71.5}{1.9} & \scell{54.0}{7.1} & \scell{79.0}{5.3} & \scell{78.0}{5.7} & \scell{47.0}{5.3} & \scell{78.0}{2.8} & \scell{73.5}{3.4} & \sbest{92.0}{5.9} \\
 & Reacher-Hazard & \scell{57.0}{2.0} & \scell{69.0}{6.2} & \scell{73.0}{7.4} & \scell{54.0}{2.8} & \scell{66.5}{4.1} & \scell{63.0}{6.2} & \scell{53.0}{4.2} & \scell{79.5}{7.9} & \scell{76.0}{6.7} & \sbest{90.5}{7.2} \\
\midrule
\multirow{3}{*}{Manipulation}
 & PushT-Pit      & \scell{37.5}{3.4} & \scell{14.5}{5.0} & \scell{12.5}{6.2} & \scell{20.5}{4.4} & \scell{15.0}{3.5} & \scell{10.5}{7.0} & \sbest{42.0}{2.8} & \scell{31.0}{5.8} & \scell{30.5}{1.9} & \scell{13.5}{4.4} \\
 & PushT-Pusher   & \scell{46.0}{2.3} & \scell{36.0}{2.8} & \scell{38.0}{7.7} & \scell{15.5}{4.1} & \scell{44.5}{6.4} & \scell{36.0}{4.3} & \sbest{65.5}{1.0} & \scell{50.5}{6.0} & \scell{54.5}{9.6} & \scell{50.5}{5.4} \\
 & Cube-Drop      & \scell{0.0}{0.0} & \scell{0.0}{0.0} & \scell{1.5}{1.0} & \scell{0.5}{1.0} & \scell{1.5}{1.9} & \scell{0.0}{0.0} & \scell{25.0}{4.8} & \sbest{32.5}{10.0} & \scell{28.5}{5.5} & \scell{0.5}{1.0} \\
\cmidrule(lr){1-12}
Average & & 45.4 & 53.6 & 57.9 & 44.2 & 60.0 & 58.1 & 40.9 & 56.9 & 49.1 & \sbestavg{66.8} \\
\bottomrule
\end{tabular}}

\normalfont\normalsize
\vspace{4pt}
\caption{\textbf{Ablation of the three components.} We report success rate (\%) and standard deviation over four seeds for all eight combinations of the components. Without the sampler, the planner uses one predicted future. Without the time-to-goal, it uses LeWM's latent distance. Without branching, it plans one action sequence. Best results are marked in \textcolor{bestred}{\textbf{RED}} and highlighted.}
\label{tab:components}
\footnotesize\ttfamily
\setlength{\tabcolsep}{0pt}
\begin{tabular*}{\textwidth}{@{\extracolsep{\fill}}l cccccccc@{}}
\toprule
 & None & \multicolumn{3}{c}{One component} & \multicolumn{3}{c}{Two components} & \textcolor{bestred}{\textbf{CoLaP}} \\
\cmidrule(lr){2-2}\cmidrule(lr){3-5}\cmidrule(lr){6-8}\cmidrule(lr){9-9}
\# & 1 & 2 & 3 & 4 & 5 & 6 & 7 & 8 \\
Sampler          &  &  & \checkmark &  & \checkmark & \checkmark &  & \checkmark \\
Time-to-goal $d$ &  & \checkmark &  &  & \checkmark &  & \checkmark & \checkmark \\
Branching        &  &  &  & \checkmark &  & \checkmark & \checkmark & \checkmark \\
\midrule
TwoRoom-Trap    & \scell{74.0}{3.7} & \scell{77.5}{5.7} & \scell{86.5}{3.4} & \scell{78.0}{7.5} & \scell{97.5}{2.5} & \scell{84.5}{4.4} & \scell{73.0}{7.0} & \sbest{99.5}{1.0} \\
TwoRoom-Bridge  & \scell{37.5}{10.4} & \scell{57.5}{2.5} & \scell{58.5}{5.3} & \scell{47.0}{7.4} & \scell{79.5}{4.7} & \scell{66.0}{3.7} & \scell{63.0}{7.4} & \sbest{83.5}{1.9} \\
Reacher-Thin    & \scell{54.0}{4.3} & \scell{67.5}{2.5} & \scell{58.0}{5.7} & \scell{60.5}{4.1} & \scell{89.0}{5.3} & \scell{53.5}{5.5} & \scell{71.5}{3.4} & \sbest{92.0}{5.9} \\
Reacher-Hazard  & \scell{56.0}{2.8} & \scell{69.0}{6.2} & \scell{44.5}{1.0} & \scell{59.5}{4.4} & \scell{89.5}{6.6} & \scell{49.5}{4.1} & \scell{73.0}{7.6} & \sbest{90.5}{7.2} \\
\cmidrule(lr){1-9}
Average          & 55.4 & 67.9 & 61.9 & 61.2 & 88.9 & 63.4 & 70.1 & \sbestavg{91.4} \\
\bottomrule
\end{tabular*}
\end{table*}

\subsection{Does CoLaP reach the goal under irreversible outcomes?}
\label{sec:exp-main}


\paragraph{Tasks.}
We experiment on nine tasks built on the four environments of LeWM~\cite{maes2026lewm,maes2026swm} (\cref{fig:tasks}). The agent is a point that moves between two rooms, a two-link arm, a pusher that moves a T-shaped block~\cite{chi2023diffusionpolicy}, or a robot arm that carries a cube~\cite{park2025ogbench}. 
Seven tasks add a random, irreversible outcome. The agent can fall through ice (TwoRoom-Trap) or through a cracked floor at one of two doors (TwoRoom-Bridge). The arm can break on a patch at the edge of its reach (Reacher-Hazard, and Reacher-Thin with a thinner patch). The block or the pusher can fall into a pit (PushT-Pit, PushT-Pusher), and the cube can slip into a hole in the table (Cube-Drop). 
A typical crossing of the hazard fails half the time, and the goal can then no longer be reached. The last two tasks test other kinds of randomness. In TwoRoom-LockedDoor, one of two doors is locked at random, and a switch shows which (\cref{ex:door}). In TwoRoom-Slip, random pushes move the agent, but nothing is irreversible, so it serves as a control.

\paragraph{Protocol.}
We follow the test protocol of LeWM~\cite{maes2026lewm}. As in LeWM, a test task pairs a start observation from a held-out episode with the goal image 25 steps later in the same episode.
We choose tasks whose straight route crosses the hazard. On TwoRoom-LockedDoor and TwoRoom-Slip, the agent starts in one room, and the goal lies in the other. For evaluation, we report the average success rate over 50 episodes on four seeds. Unless noted, planners run CEM with 300 samples and 15 iterations. They plan $H = 5$ action chunks ahead ($10$ on TwoRoom-LockedDoor) and execute the first two before they replan. 


Our goal here is to determine whether scoring a plan by its expected time to the goal helps when outcomes are irreversible. From \cref{tab:main}, CoLaP scores an average success rate of 66.8\% and is ahead of the second-best sampler in open loop by 6.8 points (\textbf{+11\% relative}). Against LeWM's published planner on the same world model, the margin is 21.4 points (\textbf{+47\% relative}). The largest gains are on the navigation and reaching tasks, where CoLaP is the best or tied for best on all six and leads by up to 13.0 points. On TwoRoom-Trap, every planner that samples several futures reaches at least 96.5\%, against 77.5\% for one predicted future. So, sampled futures show the irreversible outcome that one predicted future averages away. On the control task TwoRoom-Slip, where nothing is irreversible, CoLaP still reaches 100\%.


\begin{figure*}[t]
\centering
\includegraphics[width=\textwidth]{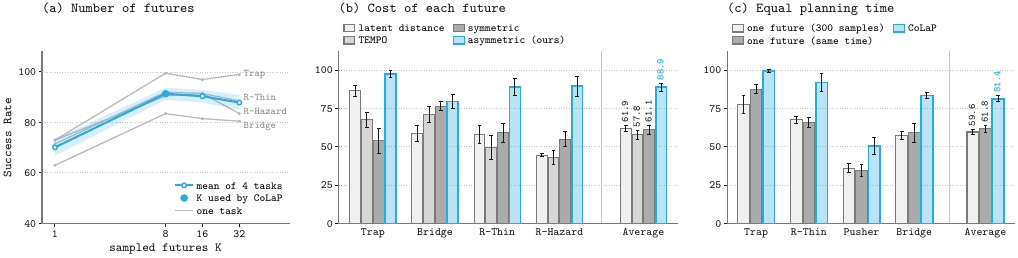}
\caption{\textbf{Design choices.} (a) Success against the number of sampled futures $K$. The thick line is the mean over four tasks, with four seed standard deviations. (b) The sampler in open loop with 8 futures, scored by four different costs. Ours is the asymmetric time-to-goal $d$. (c) One predicted future with 300 CEM samples, and with 20 to 22 times more samples so that one re-plan takes as long as one of CoLaP. R- stands for Reacher.}
\label{fig:design}
\end{figure*}

The exceptions are the three manipulation tasks, where a GCRL policy finishes ahead. In these tasks, the planner fails even without the risk. On Cube-Drop, the arm takes up most of the image, and LeWM's embedding barely changes with the positional change of the cube. So, the planner drops the cube and matches the arm pose of the goal image, as LeWM's own cube model does~\cite{maes2026lewm}. On PushT-Pit, the safe route pushes the block around the pit, and it first takes the pusher away from the goal image. 
Since the planner looks only five action chunks ahead, it rarely completes this route. So, avoiding the pit does not pay.
LeWM's published planner pushes straight across and is the best planner there.


\subsection{Why is each component needed?}
\label{sec:exp-components}
We run the ablations on TwoRoom-Trap, TwoRoom-Bridge, Reacher-Thin and Reacher-Hazard. For these tasks, if the hazard is switched off, one predicted future solves at least 99.5\% of its test tasks. So, these tasks separate the irreversible outcome from other planning failures. \Cref{tab:components} reports all eight combinations of the three components.

\paragraph{Sampled futures.}
The sampler $g_{\psi}$ turns each prediction of LeWM into a sample of the next embedding (\cref{eq:sampler}). Without it, CoLaP plans over one predicted future, and success drops on every task, by 17.5 to 26.5 points (\#8 to \#7). 
The gain of the sampler also depends on how the futures are scored. With the time-to-goal (\#2 to \#5), the sampler adds 20.0 to 22.0 points on every task. With LeWM's latent distance (\#1 to \#3), it adds only 6.5 points on average. So, the sampler shows the risk, and the time-to-goal penalizes it.

\paragraph{Time-to-goal.}
The time-to-goal $d$ scores each future by the number of action chunks it needs to reach the goal (\cref{eq:ttg}). Without it, CoLaP scores the futures with LeWM's latent distance, and success drops on every task, by 15.0 to 41.0 points (\#8 to \#6). On average, this is the largest drop of the three components. Time-to-goal is also the strongest component on its own. With one predicted future, it raises success on every task by 3.5 to 20.0 points (\#1 to \#2). So, the planner gains the most from scoring how soon a future reaches the goal, not how close it ends.


\paragraph{Branching.}
Branching lets each plan change its later actions based on what the agent observes after one or two action chunks (\cref{eq:branch}). Without it, CoLaP plans one action sequence, and success drops by only 1.0 to 4.0 points (\#8 to \#5). On these tasks, a fall comes without warning, so no observation can change the best plan. Branching pays off when an observation does change it, as with the switch in TwoRoom-LockedDoor (\cref{ex:door}). With the true simulator and the true time to the goal, branching raises success there from 59.0\% to 87.0\% (appendix). So, branching pays off when looking first saves time later.



\subsection{Which design choices matter?}
\label{sec:exp-design}

\paragraph{Number of futures.}
CoLaP scores a plan by the average time-to-goal over $K$ sampled futures (\cref{eq:ttg}). With few futures, this average is noisy. Although more futures reduce this noise, each one adds a rollout to every candidate plan.
\Cref{fig:design}a shows that success rises from 70.1\% with one predicted future to 91.4\% with eight sampled futures. Beyond eight, it no longer improves, with 90.4\% at 16 and 87.9\% at 32. Since more futures add cost without adding success, we use $K = 8$.


\paragraph{Asymmetry.}

Entering an irreversible outcome can take one action chunk, but leaving it is impossible. A symmetric time-to-goal cannot express this, so it places the agent at most one chunk farther from the goal than before the outcome (\cref{eq:bound}). To test this, \cref{fig:design}b scores the same eight sampled futures with four costs, without branching, so that only the cost changes. The asymmetric time-to-goal $d$ averages 88.9\% and is the best cost on every task. A symmetric time-to-goal reaches 61.1\%, no better than LeWM's latent distance at 61.9\%, and TEMPO's temporal distance reaches 57.8\%. 
So, only an asymmetric $d$ charges an irreversible outcome as a goal that can no longer be reached.


\paragraph{Equal planning time.}
CoLaP rolls out several futures for every candidate plan, so each re-plan takes longer. Its gain could then come from the extra compute, not from the futures. To test this, \cref{fig:design}c gives one predicted future 20 to 22 times more CEM samples, so that one re-plan takes as long as one of CoLaP. 
The extra search helps only on TwoRoom-Trap, by 10.0 points, and leaves the other tasks almost unchanged. CoLaP stays ahead on all four tasks, by 12.0 to 26.0 points. 
So, a larger search cannot recover an outcome that one prediction has already averaged away.
Further experiments and analysis are reported in Appendix.


\section{Conclusion}
\label{sec:conclusion}

We presented \textbf{Contingent Latent Planning (CoLaP)}, a planner for latent world models under stochastic irreversible outcomes. 
Latent planners score a plan by one predicted outcome and its latent distance to the goal, as if the environment were deterministic. 
So, a plan that may end in an irreversible failure can look as good as a safe one.
CoLaP instead scores each plan by its expected time to the goal over several sampled futures. 
A learned asymmetric time-to-goal makes irreversible failures costly, and the plan can branch on what the agent will observe. 
We keep the world model frozen, and CoLaP learns from offline data without rewards.
Across nine stochastic navigation, reaching, and manipulation tasks, CoLaP achieves the highest average success rate.



\paragraph{Limitations.} CoLaP changes how a plan is scored, not the world model it plans with. When planning fails even without the risk, a better score cannot help. Branching also depends on the quality of the sampled futures. 


\subsection*{LLM Usage Statement}

We used Claude Opus 5.5, a frontier model, to edit the prose of this paper. Its role was to tighten wording and to shorten passages that ran long. We also used it to write the code. The research question, the method, the proofs, and the experiments are our own, and no part of them was produced by a generative AI tool. We read every edit Claude suggested, and we take responsibility for the final content of this paper.

{
    \small
    \bibliographystyle{ieeenat_fullname}
    \bibliography{main}
}


\clearpage
\setcounter{equation}{0}
\setcounter{figure}{0}
\setcounter{table}{0}
\setcounter{algorithm}{0}
\renewcommand{\theequation}{S\arabic{equation}}
\renewcommand{\thefigure}{S\arabic{figure}}
\renewcommand{\thetable}{S\arabic{table}}
\renewcommand{\thealgorithm}{S\arabic{algorithm}}
\renewcommand{\theHequation}{S\arabic{equation}}
\renewcommand{\theHfigure}{S\arabic{figure}}
\renewcommand{\theHtable}{S\arabic{table}}
\providecommand{\theHalgorithm}{}
\renewcommand{\theHalgorithm}{S\arabic{algorithm}}
\makeatletter
\setlength{\@fptop}{0pt}
\makeatother
\maketitlesupplementary

\setcounter{tocdepth}{2}
\tableofcontents

Equations, figures, tables and algorithms numbered S1, S2, \dots{} belong to
this supplementary. Plain numbers refer to the main paper.

\appendix

\FloatBarrier
\section{Theory}
\label{app:theory}
In this section we define the theoretical framework behind the three failures of \cref{main-sec:problem} and the components that address them. \Cref{thm:meantrap} shows that a planner that sees only the mean future cannot avoid an irreversible outcome. \Cref{prop:ttg} shows what the time-to-goal learns, and \Cref{prop:wayback} shows why it must be asymmetric. \Cref{prop:info} shows when branching can help. 

\subsection{Setting, assumptions and definitions}
\label{app:theory-setting}
We use a finite set of states and identify each state with its embedding $\boldsymbol{z}$. As in \cref{main-sec:prelims}, $\boldsymbol{z}$ also holds the previous embeddings that LeWM reads. 
So, the next embedding depends only on $\boldsymbol{z}$ and the action chunk, $\boldsymbol{z}_{t+1} \sim P(\cdot \mid \boldsymbol{z}_t, \boldsymbol{a}_t)$. Once the agent reaches the goal $\boldsymbol{z}_g$, the episode ends. We judge a policy $\pi$ by the expected time of \cref{main-eq:objective}. 
Its suboptimality gap\footnote{\textbf{Suboptimality gap:} The expected time a policy loses against the best policy in the same environment.} in an environment $M$, is the expected time it loses against the best policy,
\begin{equation}
  G_M(\pi) = \mathbb{E}_{M, \pi}[\tau] - \min_{\pi'} \mathbb{E}_{M, \pi'}[\tau].
  \label{eq:supp-regret}
\end{equation}

The gap of a planner is the gap of the policy it follows. LeWM is trained with squared error, so its predictor approaches the mean
predictor~\cite{bishop2006pattern}
\begin{equation}
  m(\boldsymbol{z}, \boldsymbol{a}) = \mathbb{E}\big[\boldsymbol{z}_{t+1} \mid
  \boldsymbol{z}_t = \boldsymbol{z},\, \boldsymbol{a}_t = \boldsymbol{a}\big].
  \label{eq:supp-mean}
\end{equation}
We call a planner \emph{mean-only} if the only thing it knows about the dynamics is $m$. Its cost, search budget, and re-planning can be anything. Here, the planner of LeWM (\cref{main-eq:lewm-plan}) is mean-only, since its predictor learns $m$.

Let $\mathcal{T}$ be the set of one action-chunk transitions $(\boldsymbol{z}, \boldsymbol{z}')$ in the data, and let $n(\boldsymbol{z}, \boldsymbol{z}')$ be the fewest transitions of $\mathcal{T}$ that lead from $\boldsymbol{z}$ to $\boldsymbol{z}'$. 
Here, $n = 0$ if $\boldsymbol{z} = \boldsymbol{z}'$ and $n = \infty$ if no path exists. For an upper bound $C > 0$, we define
\begin{equation}
  d^\star(\boldsymbol{z}, \boldsymbol{z}') = \min\big(n(\boldsymbol{z}, \boldsymbol{z}'),\,
  C\big).
  \label{eq:supp-dstar}
\end{equation}
An \emph{irreversible outcome} is a set $A$ of states that the agent cannot leave and that does not contain the goal. Once the agent enters $A$, it can never reach the goal, so $\tau = L$. The fall through the ice in \Cref{main-ex:ice} is one such example. We use \textbf{four} idealized conditions.
\begin{itemize}[leftmargin=*, itemsep=1pt, topsep=2pt]
  \item[(I1)] \textbf{Exact sampler.} Every sample $\tilde{\boldsymbol{z}}^{(k)}_{t+1}$ of
  \cref{main-eq:sampler} is drawn from $P(\cdot \mid \tilde{\boldsymbol{z}}^{(k)}_t,
  \boldsymbol{a}_t)$.
  \item[(I2)] \textbf{Exact time-to-goal.} $d = d^\star$ with $C \ge L$, and $\mathcal{T}$ holds every transition that can happen.
  \item[(I3)] \textbf{Short steps.} $d(\tilde{\boldsymbol{z}}^{(k)}_t,
  \tilde{\boldsymbol{z}}^{(k)}_{t+1}) \le 1$ for every chunk of every scored future.
  \item[(I4)] \textbf{Many futures.} $K \to \infty$, so $V$ and $V_b$ equal their
  expectations.
\end{itemize}
Each result states the conditions it uses. Together, (I1) and (I2) imply (I3), since each sampled chunk is then a transition in $\mathcal{T}$, where $d^\star \le 1$.


\subsection{The mean trap}
\label{app:theory-meantrap}
LeWM learns the mean predictor $m$ of \cref{eq:supp-mean} (\cref{main-sec:problem}). Two environments can have the same $m$ even if only one of them has an irreversible outcome (\cref{main-fig:components}a). A mean-only planner gets the same inputs in both. So, it acts the same way in both, and it loses time in at least one.

\begin{theorem}[The mean trap]
\label{thm:meantrap}
Let $L$ be the episode length and $H$ the planning horizon, both in chunks. Let $T$ be an integer with $H < T < L$, and let $p \in (0, 1)$. There are two environments $M_1$ and $M_2$ with the same mean predictor $m$. In both, crossing reaches the goal in $T$ chunks. In $M_2$ only, the crossing enters an irreversible outcome with probability $p$. Every mean-only planner policy $\Pi$ has
\begin{equation}
  G_{M_1}(\Pi) + G_{M_2}(\Pi) \ge \big\lfloor p\,(L - T)/2 \big\rfloor.
  \label{eq:supp-meantrap}
\end{equation}
Under (I1), (I2) and (I4), the planner that minimizes $V$ of \cref{main-eq:ttg} has a zero gap in both.
\end{theorem}


\begin{proof}
\textit{Construction.} Both environments have the same states. At the start $\boldsymbol{z}_0$, the agent picks one of two action chunks, cross $\boldsymbol{a}_c$ or detour $\boldsymbol{a}_d$. The two routes do not meet before the goal, and neither returns to $\boldsymbol{z}_0$. The detour is deterministic and takes $T + \Delta$ chunks, where $\Delta = \lfloor p\,(L - T)/2 \rfloor$. The environments differ only in where $\boldsymbol{a}_c$ leads.
\begin{itemize}[leftmargin=*, itemsep=1pt, topsep=2pt]
  \item In $M_1$, it leads to a state with embedding $\bar{\boldsymbol{z}} = (1 - p)\,\boldsymbol{z}_c + p\,\boldsymbol{z}_f$.
  \item In $M_2$, it leads to $\boldsymbol{z}_c$ with probability $1 - p$ and to $\boldsymbol{z}_f$ with probability $p$. No transition leaves $\boldsymbol{z}_f$, so $\{\boldsymbol{z}_f\}$ is an irreversible outcome.
\end{itemize}
From $\bar{\boldsymbol{z}}$ and from $\boldsymbol{z}_c$, the goal is $T - 1$ chunks away. All other transitions are deterministic and the same in both environments.

\textit{Expected times.} The crossing takes $T$ chunks in $M_1$ and $(1 - p)\,T + p\,L = T + p\,(L - T)$ chunks in $M_2$. The detour takes $T + \Delta < L$ chunks in both. So, in $M_1$ the crossing is optimal and the detour loses $\Delta$. In $M_2$ the detour is optimal and the crossing loses $p\,(L - T) - \Delta \ge \Delta$.

\textit{Same mean.} In $M_2$, the next embedding after $\boldsymbol{a}_c$ has mean $(1 - p)\,\boldsymbol{z}_c + p\,\boldsymbol{z}_f = \bar{\boldsymbol{z}}$, as in $M_1$. All other transitions agree, so $m$ is the same in both. Before its first chunk, $\Pi$ therefore gets the same inputs in both environments, so it picks $\boldsymbol{a}_c$ with the same probability $q$ in both. The first chunk fixes the route, since the routes do not meet. So, $G_{M_1}(\Pi) \ge (1 - q)\,\Delta$ and $G_{M_2}(\Pi) \ge q\,\Delta$, and the two add up to at least $(1 - q)\,\Delta + q\,\Delta = \Delta$.

\textit{Minimizing $V$.} By (I1) and (I4), $V$ is the expectation of $J^{(k)}$ under the true transitions. By (I2), $d^\star(\boldsymbol{z}, \boldsymbol{z}_g)$ is the fewest chunks from $\boldsymbol{z}$ to the goal, and $d^\star(\boldsymbol{z}_f, \boldsymbol{z}_g) = C \ge L$, since no path leaves irreversible outcome $\boldsymbol{z}_f$. Since $H < T$, no future reaches the goal within the horizon, so $J^{(k)} = H + d^\star(\tilde{\boldsymbol{z}}^{(k)}_{H+1}, \boldsymbol{z}_g)$. The best plan along the detour has $V = T + \Delta$. The best plan along the crossing has $V = T$ in $M_1$, and $V = (1 - p)\,T + p\,(H + C) > T + p\,(L - T)$ in $M_2$, since $H + C > L$. So, the first chunk starts an optimal route in both environments. After it, all transitions are deterministic and $d^\star$ is exact, so each replan follows an optimal path, and the gap is zero.
\end{proof}


\subsection{What the time-to-goal learns}
\label{app:theory-ttg}
The time-to-goal $d$ is trained with QRL (\cref{main-eq:qrl}). QRL shows that this objective recovers the optimal time to the goal when the dynamics are deterministic~\cite{wang2023qrl}. Here, we state what the same objective learns from data with random and irreversible outcomes. We leave out the noise constraint of \cref{main-eq:noise}, which acts on embeddings off the data.

\begin{proposition}[What the time-to-goal learns]
\label{prop:ttg}
Consider \cref{main-eq:qrl} with $\eta = 0$. Its constraint then reads $d(\boldsymbol{z}, \boldsymbol{z}') \le 1$ for every $(\boldsymbol{z}, \boldsymbol{z}') \in \mathcal{T}$, and we call $d$ \emph{feasible} if it meets it. Let $\varphi$ of \cref{main-eq:qrl} be strictly increasing, and let $\rho$ be the distribution of the pairs $(\boldsymbol{z}, \boldsymbol{z}')$ in the objective. Let $d$ range over all quasimetrics with values in $[0, C]$, that is, all such $d$ with $d(\boldsymbol{z}, \boldsymbol{z}) = 0$ that obey the triangle inequality. Then
\begin{itemize}[leftmargin=*, itemsep=1pt, topsep=2pt]
  \item[(i)] $d^\star$ is feasible,
  \item[(ii)] every feasible $d$ has $d \le d^\star$ on all pairs, and
  \item[(iii)] every maximizer equals $d^\star$ on every pair with $\rho(\boldsymbol{z}, \boldsymbol{z}') > 0$.
\end{itemize}
\end{proposition}

\begin{proof}
\textit{(i)} $d^\star$ has values in $[0, C]$ and $d^\star(\boldsymbol{z}, \boldsymbol{z}) = 0$. A path from $\boldsymbol{z}$ to $\boldsymbol{z}'$ followed by a path from $\boldsymbol{z}'$ to $\boldsymbol{z}''$ is a path from $\boldsymbol{z}$ to $\boldsymbol{z}''$, so $n(\boldsymbol{z}, \boldsymbol{z}'') \le n(\boldsymbol{z}, \boldsymbol{z}') + n(\boldsymbol{z}', \boldsymbol{z}'')$. The upper bound $C$ keeps this inequality, since $\min(a + b, C) \le \min(a, C) + \min(b, C)$ for $a, b \ge 0$. So, $d^\star$ is a quasimetric. Every transition in $\mathcal{T}$ is a path of length $1$, so $d^\star \le 1$ on $\mathcal{T}$.

\textit{(ii)} Consider a feasible $d$ and a pair $(\boldsymbol{z}, \boldsymbol{z}')$. If $n(\boldsymbol{z}, \boldsymbol{z}') = k < \infty$, take a path $\boldsymbol{z} = \boldsymbol{x}_0, \boldsymbol{x}_1, \dots, \boldsymbol{x}_k = \boldsymbol{z}'$ of $k$ transitions in $\mathcal{T}$. The triangle inequality and feasibility give $d(\boldsymbol{z}, \boldsymbol{z}') \le \sum_{i=0}^{k-1} d(\boldsymbol{x}_i, \boldsymbol{x}_{i+1}) \le k$. Since also $d \le C$, we get $d(\boldsymbol{z}, \boldsymbol{z}') \le \min(k, C) = d^\star(\boldsymbol{z}, \boldsymbol{z}')$. If $n = \infty$, then $d^\star = C \ge d$.

\textit{(iii)} By (ii) and since $\varphi$ is increasing, $\mathbb{E}_\rho[\varphi(d)] \le \mathbb{E}_\rho[\varphi(d^\star)]$ for every feasible $d$. So, by (i), $d^\star$ is a maximizer. A maximizer $d$ reaches equality, which needs $\varphi(d) = \varphi(d^\star)$ on every pair with $\rho > 0$. Since $\varphi$ is strictly increasing, $d = d^\star$ there.
\end{proof}


So, $d$ learns the fewest chunks between two states in the data, up to the upper bound $C$ (\cref{main-sec:ttg}). In our implementation, $\varphi$ flattens above $C$ (\cref{app:impl-ttg}), so $C$ acts as a soft upper bound. \Cref{prop:ttg} describes the exact maximizer over all quasimetrics, which the network only approximates. Two episodes rarely share an exact embedding. So, paths across episodes come from the generalization of the network.


\begin{corollary}[Irreversibility without labels]
\label{cor:irrev}
Let $A$ be a set of states with $\boldsymbol{z}_g \notin A$, such that no transition in $\mathcal{T}$ leaves $A$. Then
\begin{itemize}[leftmargin=*, itemsep=1pt, topsep=2pt]
  \item[(i)] $d^\star(\boldsymbol{z}', \boldsymbol{z}) = C$ for every $\boldsymbol{z}' \in A$ and $\boldsymbol{z} \notin A$. In particular, $d^\star(\boldsymbol{z}', \boldsymbol{z}_g) = C$.
  \item[(ii)] $d^\star(\boldsymbol{z}, \boldsymbol{z}') \le 1$ for every $(\boldsymbol{z}, \boldsymbol{z}') \in \mathcal{T}$ with $\boldsymbol{z}' \in A$.
\end{itemize}
\end{corollary}

\begin{proof}
\textit{(i)} Every path in $\mathcal{T}$ that starts in $A$ stays in $A$, so it never reaches $\boldsymbol{z} \notin A$. So, $n(\boldsymbol{z}', \boldsymbol{z}) = \infty$ and $d^\star(\boldsymbol{z}', \boldsymbol{z}) = C$. \textit{(ii)} follows from \Cref{prop:ttg}(i).
\end{proof}


An irreversible outcome of the environment has no transition out of it, and neither does the data. So, $d$ prices the way into it at no more than one chunk and the way out at the upper bound $C$. This is how $d$ learns which outcomes are irreversible from the data alone, without failure labels (\cref{main-sec:ttg})~\cite{grinsztajn2021there}. 


\subsection{An irreversible outcome costs at most the way back}
\label{app:theory-wayback}

\Cref{main-eq:bound} shows that, with a symmetric $d$, entering an irreversible outcome adds at most one chunk to the time-to-goal. Here, we bound how much it can add to the score $J^{(k)}$, for any quasimetric $d$.

\begin{proposition}[An irreversible outcome costs at most the way back]
\label{prop:wayback}
Let $d$ be a quasimetric that meets (I3). Consider two futures $k$ and $k'$ of the same plan that agree up to chunk $s$, with $1 \le s \le H$, so that $\tilde{\boldsymbol{z}}^{(k)}_t = \tilde{\boldsymbol{z}}^{(k')}_t = \tilde{\boldsymbol{z}}_t$ for $t \le s$. Let $B = d(\tilde{\boldsymbol{z}}^{(k)}_{H+1}, \tilde{\boldsymbol{z}}_s)$ be the way back from the end of future $k$ to their last common embedding. Then
\begin{equation}
  J^{(k)} - J^{(k')} \le 2\,(H + 1 - s) + B.
  \label{eq:supp-wayback}
\end{equation}
\end{proposition}


\begin{proof}
The running terms of $J^{(k)}$ and $J^{(k')}$ are equal for $t < s$. The other $H + 1 - s$ terms lie in $[0, 1]$, so they add at most $H + 1 - s$ to $J^{(k)} - J^{(k')}$. For the last term, we use the triangle inequality, and then (I3) on the $H + 1 - s$ chunks of future $k'$ after $\tilde{\boldsymbol{z}}_s$,
\begin{equation}
  \begin{aligned}
    d(\tilde{\boldsymbol{z}}^{(k)}_{H+1}, \boldsymbol{z}_g)
    &\le B + d(\tilde{\boldsymbol{z}}_s, \boldsymbol{z}_g)\\
    &\le B + (H + 1 - s) + d(\tilde{\boldsymbol{z}}^{(k')}_{H+1}, \boldsymbol{z}_g).
  \end{aligned}
  \label{eq:supp-wayback-proof}
\end{equation}
Adding the two bounds gives the claim.
\end{proof}

\begin{corollary}[A symmetric time-to-goal undercharges an irreversible outcome]
\label{cor:sym}
If $d$ in \Cref{prop:wayback} is also symmetric, then $J^{(k)} - J^{(k')} \le 3\,(H + 1 - s) \le 3H$, for every episode length $L$.
\end{corollary}

\begin{proof}
By symmetry, $B = d(\tilde{\boldsymbol{z}}_s, \tilde{\boldsymbol{z}}^{(k)}_{H+1})$. (I3) on the $H + 1 - s$ chunks of future $k$ after $\tilde{\boldsymbol{z}}_s$ gives $B \le H + 1 - s$.
\end{proof}


So, with a symmetric $d$, entering an irreversible outcome adds at most $3H$ chunks to the score of a future, however long the episode is. A planner can then prefer a route that risks an irreversible outcome over a safe route that takes longer. Under $d^\star$, the way back is $B = C \ge L$ (\Cref{cor:irrev}), so entering an irreversible outcome costs the whole episode. 

\subsection{Information must come before the irreversible outcome}
\label{app:theory-info}
Branching lets each group of futures keep the root plan or switch to the alternative after chunk $b$ (\cref{main-eq:branch}). Here, we show when this lowers the value of a plan.


\begin{proposition}[Information must come first]
\label{prop:info}
Fix a candidate pair and a branch chunk $b$. We call $\boldsymbol{a}_{b+1:H}$ and $\boldsymbol{a}'_{b+1:H}$ its two continuations. Let $V^{\mathrm{ol}} = \min\{V(\boldsymbol{a}_{1:H}),\, V(\boldsymbol{a}_{1:b}, \boldsymbol{a}'_{b+1:H})\}$ be the value of the two single action sequences.
\begin{enumerate}[label=(\roman*), leftmargin=*, itemsep=1pt, topsep=2pt]
  \item $V_b \le V^{\mathrm{ol}}$, with equality if one continuation is best for both groups.
  \item Under (I1), (I2) and (I4), let every future at chunk $b + 1$ be either in an irreversible outcome $A$ or at one common embedding $\bar{\boldsymbol{z}}$. Then $V_b = V^{\mathrm{ol}}$.
\end{enumerate}
\end{proposition}

\begin{proof}
\textit{(i)} For a continuation $\boldsymbol{c}$, let $S_j(\boldsymbol{c}) = \sum_{k \in \mathcal{G}_j} J^{(k)}(\boldsymbol{a}_{1:b}, \boldsymbol{c})$ be the total score of group $j$. Then
\begin{equation}
  \begin{aligned}
    K\,V_b &= \min_{\boldsymbol{c}} S_1(\boldsymbol{c}) + \min_{\boldsymbol{c}} S_2(\boldsymbol{c}),\\
    K\,V(\boldsymbol{a}_{1:b}, \boldsymbol{c}) &= S_1(\boldsymbol{c}) + S_2(\boldsymbol{c}).
  \end{aligned}
  \label{eq:supp-info-proof}
\end{equation}
The first line is at most the second for every $\boldsymbol{c}$, so $V_b \le V^{\mathrm{ol}}$. If one $\boldsymbol{c}$ minimizes both sums, then $V_b = V(\boldsymbol{a}_{1:b}, \boldsymbol{c}) \ge V^{\mathrm{ol}}$, so $V_b = V^{\mathrm{ol}}$.

\textit{(ii)} Consider a future in $A$. By (I1), it stays in $A$, and by (I2) and \Cref{cor:irrev}, $d^\star = C \ge 1$ from $A$ to the goal. So, its running terms after chunk $b$ are $1$ and its last term is $C$, and its $J^{(k)}$ is the same for both continuations. The other futures are at $\bar{\boldsymbol{z}}$. By (I1), they follow the same law after chunk $b$, so by (I4), each adds the same expected $J^{(k)}$ to its group for a given $\boldsymbol{c}$. So, the same $\boldsymbol{c}$ minimizes $S_1$ and $S_2$, and (i) gives $V_b = V^{\mathrm{ol}}$.
\end{proof}


So, branching lowers the value only if the agent observes something before an irreversible outcome that changes its best continuation. This is the classical value of information~\cite{howard1966information,kaelbling1998planning}, applied to contingent plans in latent space. With finite $K$, (i) still holds for every sample, so $V^{\mathrm{ol}} - V_b \ge 0$, and noise alone can make it positive. Also, $V_b$ is the value of a policy that switches after chunk $b$. The agent replans after two chunks, which matches to $b = 2$. 

\FloatBarrier
\section{Algorithm}
\label{app:algorithm}

\Cref{alg:supp-colap} states how CoLaP plans (\cref{main-sec:objective}). At every
re-plan, CEM searches over pairs of a root plan $\boldsymbol{a}_{1:H}$ and an alternative plan $\boldsymbol{a}'_{1:H}$ (\cref{main-sec:contingency}). It scores each candidate pair in three steps. The sampler rolls out $K$ futures of both plans, the time-to-goal $d$ scores every future, and the futures are grouped by what the agent will observe, so that each group keeps the root or switches to the alternative. The agent then executes the first two chunks of the best plan and then replans. All settings are in \cref{app:impl}.

\begin{algorithm}[!htbp]
\caption{Contingent Latent Planning (CoLaP)}
\label{alg:supp-colap}
\begin{algorithmic}[1]
  \Require Frozen LeWM ($\mathrm{enc}_{\theta}$, $\mathrm{pred}_{\phi}$), sampler $g_{\psi}$, time-to-goal $d$,
  goal image $\boldsymbol{o}_g$, horizon $H$, futures $K$
  \State Encode the goal, $\boldsymbol{z}_g \leftarrow \mathrm{enc}_{\theta}(\boldsymbol{o}_g)$
  \For{each replanning step}
    \State Encode the current observation, $\tilde{\boldsymbol{z}}^{(k)}_1 \leftarrow \mathrm{enc}_{\theta}(\boldsymbol{o})$
    \State Initialize a Gaussian over plan pairs $(\boldsymbol{a}_{1:H}, \boldsymbol{a}'_{1:H})$
    \For{each CEM iteration}
      \State Draw noise $\boldsymbol{\epsilon}^{(k)}_{1:H}$ for $k = 1, \dots, K$
      \State Sample candidate pairs from the Gaussian
      \For{each candidate, with the same noise}
        \State Roll out $K$ futures of both plans \Comment{\cref{main-eq:sampler}}
        \State Score every future with $J^{(k)}$ \Comment{\cref{main-eq:ttg}}
        \State Group the futures, $V \leftarrow \min_b V_b$ \Comment{\cref{main-eq:branch}}
      \EndFor
      \State Refit the Gaussian to the lowest-$V$ candidates
    \EndFor
    \State Execute the first two chunks of the best plan $\boldsymbol{a}^{*}_{1:H}$
  \EndFor
\end{algorithmic}
\end{algorithm}

\FloatBarrier
\section{Experimental Setup}
\label{app:setup}

\subsection{Compute}
\label{app:setup-compute}
We trained every world model and every component, and ran every evaluation, on one
workstation with an NVIDIA GeForce RTX 5090 with 32\,GB of memory, an Intel Core i9-14900K
with 24 cores, and 61\,GB of system memory, under Ubuntu 26.04. The code uses PyTorch 2.14
with CUDA 13.0 and Python 3.10, on top of stable-worldmodel~\cite{maes2026swm} at commit
\texttt{8b2e8c6}.

\subsection{Environments and datasets}
\label{app:setup-envs}

\begin{table*}[!htbp]
\centering
\caption{\textbf{Tasks.} The random event of each task, whether it is irreversible, the
source of the data, and the number of episodes used to train LeWM and the sampler. The
sampler episodes are collected the same way, and LeWM never sees them.}
\label{tab:supp-tasks}
\footnotesize
\setlength{\tabcolsep}{5pt}
\begin{tabularx}{\textwidth}{@{}llYcYl@{}}
\toprule
Environment & Task & Random event & Irreversible & Data & Episodes (LeWM + sampler) \\
\midrule
\multirow{4}{*}{TwoRoom}
 & TwoRoom-Trap       & the agent falls through the ice              & \checkmark & expert                                   & 10{,}000 + 5{,}000 \\
 & TwoRoom-Bridge     & the agent falls at the cracked door          & \checkmark & expert, takes the nearer door            & 10{,}000 + 5{,}000 \\
 & TwoRoom-Slip       & a push of 5\,px in a random direction        &            & expert                                   & 10{,}000 + 5{,}000 \\
 & TwoRoom-LockedDoor & one of two doors is locked                   &            & expert, picks a door before the lamps light & 6{,}000 + 5{,}000 \\
\midrule
\multirow{2}{*}{Reacher}
 & Reacher-Thin       & the arm breaks on a 4\,cm patch              & \checkmark & random actions, 200 steps                & 10{,}000 + 5{,}000 \\
 & Reacher-Hazard     & the arm breaks on a 7\,cm patch              & \checkmark & random actions, 200 steps                & 10{,}000 + 5{,}000 \\
\midrule
\multirow{2}{*}{PushT}
 & PushT-Pit          & the block falls into the pit                 & \checkmark & PushT expert, replayed                   & 12{,}685 + 5{,}000 \\
 & PushT-Pusher       & the pusher falls into the pit                & \checkmark & PushT expert, replayed                   & 12{,}685 + 5{,}000 \\
\midrule
Cube
 & Cube-Drop          & the cube falls into the hole                 & \checkmark & OGBench scripted expert                  & 10{,}000 + 5{,}000 \\
\bottomrule
\end{tabularx}
\end{table*}

\Cref{tab:supp-tasks} lists the nine tasks. Every task adds randomness to an environment of
LeWM~\cite{maes2026lewm,maes2026swm}.


\paragraph{Hazards.} Each of the seven hazard tasks has a fixed hazard region that is visible in every frame. At each environment step, the agent or the object it moves enters the irreversible outcome with probability $h = 1 - 0.5^{1/n}$. Here, $n$ is the number of steps of a typical crossing, between 4 and 8 depending on the task. So, a typical crossing fails with probability 0.5, as in \cref{main-ex:ice}. After the irreversible outcome, the agent or the object freezes and turns dark, and the goal can no longer be reached. 

\paragraph{TwoRoom.} A point agent moves between two rooms. On TwoRoom-Trap, it can fall through an ice patch in the left room. On TwoRoom-Bridge, the wall between the rooms has two doors. The agent can fall through the cracked floor at one door, while the other door, 36\,px lower, is safe. On TwoRoom-Slip, a push of 5\,px in one of eight directions moves the agent with probability 0.3 at every step. Nothing is irreversible in this task, as it serves as control. On TwoRoom-LockedDoor, one of the two doors is locked at random. The agent learns which one when it reaches a switch or stands next to a door. Two lamps then show the locked door for the rest of the episode. The expert of this task picks a door before the lamps light up, so its actions carry no information about the lock.

\paragraph{Reacher.} The two-link arm of the DeepMind Control Suite~\cite{tassa2018deepmind} must reach a goal pose. A patch lies at the outer edge of its reach, 7\,cm deep on Reacher-Hazard and 4\,cm deep on Reacher-Thin. The arm breaks on the patch, but it can avoid the patch by bending the wrist. As in LeWM, the data are episodes of random actions. The arm breaks in 21.6\% of them on Reacher-Hazard and in 15.5\% on Reacher-Thin.

\paragraph{PushT and Cube.} In PushT~\cite{chi2023diffusionpolicy}, a pusher moves a T-shaped block. On PushT-Pit, the block can fall into a pit, and the pusher crosses it safely. On PushT-Pusher, the pusher can fall into a 48\,px pit, and the block crosses it safely. 
The data from both tasks replay the expert episodes of PushT with the pit added. So, the replayed episodes differ from the originals only where something falls. 
The pusher falls in 23.5\% of the PushT-Pusher episodes. On Cube-Drop, the single-cube task of OGBench~\cite{park2025ogbench} has an 8\,cm hole in the table. A carried cube can fall into the hole, while the arm passes over it safely.

\FloatBarrier
\section{Implementation Details}
\label{app:impl}

\subsection{World model}
\label{app:impl-lewm}
We train one LeWM per task from scratch with stable-worldmodel~\cite{maes2026swm} and keep it frozen. The encoder is a ViT-Tiny~\cite{dosovitskiy2021image} that maps each $224 \times 224$ frame to a 192-d embedding. The predictor is a 6-layer transformer. It reads the last three embeddings and action chunks and predicts the next embedding. We use this history of three frames on every task. One action-chunk is five environment actions.

\subsection{Future sampler}
\label{app:impl-sampler}
The sampler $g_{\psi}$ is an MLP with two hidden layers of width 512. It reads the prediction $\hat{\boldsymbol{z}}$ and 16-d Gaussian noise $\boldsymbol{\epsilon}$, and returns the offset added to $\hat{\boldsymbol{z}}$ in \cref{main-eq:sampler}. Its last layer is initialized to zero, so before training it returns the prediction of LeWM. It trains on windows of four frames from the 5{,}000 held-out episodes. The frozen LeWM predicts the fourth embedding from the first three real frames, and the energy score of \cref{main-eq:es} compares 8 draws with the real fourth embedding.


\subsection{Time-to-goal}
\label{app:impl-ttg}
The time-to-goal $d$ is the MRN of \cref{main-eq:mrn}. One MLP with two hidden layers of width 512, maps a standardized embedding to the two 64-d heads $f$ and $h$. We train $d$ with \cref{main-eq:qrl,main-eq:noise} on all frames of the training episodes of LeWM. Here, a transition is two frames one action chunk apart in the same episode, and a random pair is two frames drawn independently from the whole dataset. 
We use $\varphi(d) = -\tfrac{C}{5}\,\mathrm{softplus}(5 - 5d/C)$, where $C = 50$ chunks is the upper bound $d_{\max}$ of \cref{main-sec:ttg}. Both constraints use the tolerance $\eta = 0.25$. The noise constraint uses $\delta = 0.5$, with $\sigma \sim \mathcal{U}[0, 0.2]$ drawn for each pair. Each constraint has its own Lagrange multiplier, which starts at 0.01, stays positive through a softplus, and is updated by dual ascent.

\subsection{Planner and branching}
\label{app:impl-planner}

CEM searches over the root plan and the alternative together, as one plan of $2H$ chunks (\cref{main-sec:contingency}). Each iteration samples 300 candidate pairs from a diagonal Gaussian over normalized actions and scores them with \cref{main-eq:full}. It then refits the mean and the standard deviation to the 30 pairs with the lowest value. CEM runs for 15 iterations, and the first candidate of each iteration is the current mean. The standard deviation starts at 1, and the candidates are not clipped to the action range (\cref{app:abl-clip} clips them). The agent executes the first two chunks of the best root plan and then replans. At the next re-plan, the mean starts from the chunks of the old root that were not executed, and all other chunks start at 0.

Each candidate pair is scored on $K = 8$ futures. At the branch chunk $b \in \{1, 2\}$, 2-means with 6 iterations splits the sampled embeddings $\tilde{\boldsymbol{z}}^{(k)}_{b+1}$ into the two groups of \cref{main-eq:branch}. 

\subsection{Hyperparameters}
\label{app:impl-hparams}
\Cref{tab:supp-hparams} lists the settings shared by all nine tasks, and \cref{tab:supp-pertask} lists the task specific settings. 

\begin{table}[!htbp]
\centering
\caption{\textbf{Shared hyperparameters.} These are the same on all nine tasks.}
\label{tab:supp-hparams}
\footnotesize
\setlength{\tabcolsep}{4pt}
\begin{tabularx}{\linewidth}{@{}lY@{}}
\toprule
Hyperparameter & Value \\
\midrule
\multicolumn{2}{@{}l}{\textit{World model (LeWM)}} \\
Encoder & ViT-Tiny, patch 14, $224 \times 224$ \\
Embedding & 192-d \\
Predictor & transformer, 6 layers, 16 heads of 64-d \\
Predictor MLP width, dropout & 2048, 0.1 \\
History & 3 frames \\
Chunk (frame-skip) & 5 actions \\
SIGReg weight & 0.09 (17 knots, 1{,}024 projections) \\
Optimizer & AdamW, lr $5 \times 10^{-5}$, weight decay $10^{-3}$ \\
Batch size & 128 \\
Precision, gradient clipping & bf16, 1.0 \\
\midrule
\multicolumn{2}{@{}l}{\textit{Future sampler $g_{\psi}$}} \\
Network & MLP [512, 512], SiLU, last layer at 0 \\
Noise $\boldsymbol{\epsilon}$ & 16-d Gaussian \\
Draws per window & 8 \\
Optimizer & Adam, lr $10^{-3}$ \\
Batch size, epochs & 1{,}024 windows, 20 \\
Data & 5{,}000 held-out episodes \\
Risky windows & half of every epoch \\
\midrule
\multicolumn{2}{@{}l}{\textit{Time-to-goal $d$}} \\
Network & MRN, MLP [512, 512], SiLU, two 64-d heads \\
$\varphi$ & $-\tfrac{C}{5}\,\mathrm{softplus}(5 - 5d/C)$, $C = 50$ \\
Tolerance $\eta$ & 0.25 \\
Noise constraint & $\delta = 0.5$, $\sigma \sim \mathcal{U}[0, 0.2]$ \\
Optimizer & Adam, lr $3 \times 10^{-4}$ \\
Multipliers & start 0.01, dual ascent with Adam, lr $10^{-2}$ \\
Batch size, steps & 1{,}024, 60{,}000 \\
\midrule
\multicolumn{2}{@{}l}{\textit{Planner}} \\
CEM samples, elites, iterations & 300, 30, 15 \\
Initial standard deviation & 1 \\
Horizon $H$ & 5 chunks \\
Chunks executed per re-plan & 2 \\
Futures $K$ & 8 \\
Alternative plans & 1 \\
Branch chunks $b$ & 1 and 2 \\
Groups & 2, by 2-means with 6 iterations \\
\bottomrule
\end{tabularx}
\end{table}

\begin{table}[!htbp]
\centering
\caption{\textbf{Task-specific settings.} Epochs train LeWM, the budget is the most environment steps an episode may take, and $H$ is the planning horizon in action chunks.}
\label{tab:supp-pertask}
\footnotesize
\setlength{\tabcolsep}{0pt}
\begin{tabular*}{\linewidth}{@{\extracolsep{\fill}}lccc@{}}
\toprule
Task & Epochs & Budget & $H$ \\
\midrule
TwoRoom-Trap       & 100 & 50  & 5  \\
TwoRoom-LockedDoor & 100 & 100 & 10 \\
TwoRoom-Slip       & 30  & 50  & 5  \\
TwoRoom-Bridge     & 30  & 200 & 5  \\
Reacher-Thin       & 30  & 200 & 5  \\
Reacher-Hazard     & 30  & 200 & 5  \\
PushT-Pusher       & 30  & 200 & 5  \\
PushT-Pit          & 30  & 50  & 5  \\
Cube-Drop          & 30  & 50  & 5  \\
\bottomrule
\end{tabular*}
\end{table}

\FloatBarrier
\section{Baselines}
\label{app:baselines}
We compare CoLaP with nine baselines from three families (\cref{main-sec:exp-main}). The six planners use the same frozen LeWM as CoLaP, and the three GCRL policies use its embeddings. Every baseline runs on four seeds of 50 episodes each. 

\begin{table}[!htbp]
\centering
\caption{\textbf{Planners.} All planners use the frozen LeWM and CEM with 300 candidates, refitted to the 30 with the lowest cost, over $H = 5$ chunks ($10$ on TwoRoom-LockedDoor). Iterations is the number of CEM iterations per re-plan, which for LeWM is 30 on PushT and 10 elsewhere.}
\label{tab:supp-baselines}
\footnotesize
\setlength{\tabcolsep}{3pt}
\begin{tabularx}{\linewidth}{@{}lcYcc@{}}
\toprule
Planner & Futures & Score of a plan & Iterations & Chunks \\
\midrule
LeWM               & 1 predicted   & latent distance, \cref{main-eq:lewm-plan} & 30 / 10 & 5 \\
One future         & 1 predicted   & $J$, \cref{eq:supp-onefuture}               & 15 & 2 \\
MC-dropout         & 8 dropout     & $V$, \cref{main-eq:ttg}                     & 15 & 2 \\
TEMPO              & 1 predicted   & $\mathcal{C}_{\text{TEMPO}}$, \cref{eq:supp-tempo} & 15 & 2 \\
Sampler            & 8 sampled     & $V$, \cref{main-eq:ttg}                     & 15 & 2 \\
CVaR               & 8 sampled     & $\mathrm{CVaR}_{0.25}$, \cref{eq:supp-cvar} & 15 & 2 \\
\midrule
CoLaP              & 8 sampled     & $\min_b V_b$, \cref{main-eq:full}           & 15 & 2 \\
\bottomrule
\end{tabularx}
\end{table}

\subsection{LeWM planners}
\label{app:baselines-lewm}

\paragraph{LeWM~\cite{maes2026lewm}.} This is the published planner of LeWM, so it shows what
the world model achieves as it was released. It rolls out one predicted future and scores a
plan by the distance of its last prediction to the goal,
$\mathcal{C}(\hat{\boldsymbol{z}}_{H+1}) = \lVert \hat{\boldsymbol{z}}_{H+1} - \boldsymbol{z}_g \rVert_2^2$
(\cref{main-eq:lewm-plan}). As in LeWM, CEM draws 300 candidates and keeps 30 with the lowest cost, for 30
iterations on PushT and 10 on the other environments. The plan has $H = 5$ action chunks ($10$ on
TwoRoom-LockedDoor), and the agent executes five of them before it replans.

\paragraph{One predicted future.} This planner keeps the single prediction of LeWM but scores
it with our time-to-goal. Its cost is \cref{main-eq:ttg} with $K = 1$ and the prediction in
place of a sample,
\begin{equation}
  \begin{aligned}
    J(\boldsymbol{a}_{1:H}) &= \sum_{t=1}^{H} \min\big(d(\hat{\boldsymbol{z}}_{t+1}, \boldsymbol{z}_g),\, 1\big)\\
    &\quad + d(\hat{\boldsymbol{z}}_{H+1}, \boldsymbol{z}_g).
  \end{aligned}
  \label{eq:supp-onefuture}
\end{equation}

\paragraph{MC-dropout~\cite{gal2016dropout}.} MC-dropout draws several predictions from a
network that was trained to give one. We keep the dropout of the LeWM predictor on at test time 
(rate 0.1,\cref{tab:supp-hparams}). The $k$-th future draws new dropout masks
$\boldsymbol{m}^{(k)}_t$ at every step,
\begin{equation}
  \hat{\boldsymbol{z}}^{(k)}_{t+1} = \mathrm{pred}_{\phi}\big(\hat{\boldsymbol{z}}^{(k)}_t, \boldsymbol{a}_t;\, \boldsymbol{m}^{(k)}_t\big),
  \qquad k = 1, \dots, K,
  \label{eq:supp-mcd}
\end{equation}
and the plan is scored by $V$ in \cref{main-eq:ttg} with $K = 8$. Unlike our noise, the
masks are not shared between candidates. Dropout spreads the futures by the uncertainty of
the network, not by the randomness of the environment. So, MC-dropout tests whether any
spread of futures is enough.

\paragraph{TEMPO~\cite{moukheiber2026tempo}.} TEMPO also replaces the latent distance with a
learned temporal distance, but a symmetric one. An MLP $F$ maps each embedding to a space
where distance measures time. It is trained on pairs of frames $k$ steps apart in the same
episode,
\begin{equation}
  \begin{aligned}
    \mathcal{L}_{\text{TEMPO}} \triangleq\;
    & \mathbb{E}_{k \le 75}\big[\ell_{\delta}\big(\lVert F(\boldsymbol{z}_t) - F(\boldsymbol{z}_{t+k}) \rVert_2,\; k/\kappa\big)\big]\\
    & + \tfrac{1}{2}\, \mathbb{E}_{75 < k < 200}\big[\big[3 - \lVert F(\boldsymbol{z}_t) - F(\boldsymbol{z}_{t+k}) \rVert_2\big]_+\big],
  \end{aligned}
  \label{eq:supp-tempo-loss}
\end{equation}
where $\ell_{\delta}$ is the smooth L1 loss and $\kappa = 25$ steps. TEMPO plans with a blend
of both distances at the last step (its Eq.~5),
\begin{equation}
  \begin{aligned}
    \mathcal{C}_{\text{TEMPO}} \triangleq\;
    & (1 - w)\, \lVert \hat{\boldsymbol{z}}_{H+1} - \boldsymbol{z}_g \rVert_2^2\\
    & + w\, \bar{d}^{\,2}\, \lVert F(\hat{\boldsymbol{z}}_{H+1}) - F(\boldsymbol{z}_g) \rVert_2^2,
  \end{aligned}
  \label{eq:supp-tempo}
\end{equation}
with $w = 0.5$. Here, $\bar{d}$ is the median latent distance between frames 25 steps apart
in the training data, so both terms have the same scale. We reimplement TEMPO offline with
one predicted future. $F$ is an MLP with two hidden layers of width 256 and a 64-d output,
trained for 3{,}000 steps. TEMPO also updates $F$ online, which we do not do.
Since the distance of $F$ is symmetric, the bound of \cref{main-eq:bound} holds for it. So,
TEMPO tests whether a symmetric temporal distance is enough.

\subsection{Our futures in open loop}
\label{app:baselines-ol}
These two planners use our sampler and our time-to-goal, but plan one action sequence for
all futures. Like CoLaP, they use $K = 8$ futures.

\paragraph{Sampler in open loop.} It scores a plan by $V$ in \cref{main-eq:ttg}, the mean
time-to-goal over the sampled futures.

\paragraph{CVaR~\cite{rockafellar2000optimization}.} CVaR is a standard risk-averse
alternative to the mean. It scores a plan by the mean over its worst futures,
\begin{equation}
  \mathrm{CVaR}_{\alpha}(\boldsymbol{a}_{1:H}) = \frac{1}{m} \sum_{k \in \mathcal{W}_m} J^{(k)}(\boldsymbol{a}_{1:H}),
  \qquad m = \lceil \alpha K \rceil,
  \label{eq:supp-cvar}
\end{equation}
where $\mathcal{W}_m$ holds the $m$ futures with the largest $J^{(k)}$. With $\alpha = 0.25$
and $K = 8$, it averages the worst 2 futures. CVaR weights bad futures more than their
probability, while CoLaP weights all futures equally and needs no risk level. So, CVaR tests
whether risk aversion can replace branching.

\subsection{Offline goal-conditioned RL}
\label{app:baselines-gcrl}

\begin{table}[!htbp]
\centering
\caption{\textbf{GCRL hyperparameters.}}
\label{tab:supp-gcrl}
\footnotesize
\setlength{\tabcolsep}{4pt}
\begin{tabularx}{\linewidth}{@{}lY@{}}
\toprule
Hyperparameter & Value \\
\midrule
Gradient steps & $10^6$ per stage \\
Optimizer & Adam, lr $3 \times 10^{-4}$ \\
Batch size & 1{,}024 \\
Networks & MLP, 3 layers of 512, LayerNorm, GELU \\
Input & standardized LeWM embeddings, 3 frames and 1 goal frame \\
Policy & Gaussian, state-independent log-std in $[-5, 2]$ \\
Discount $\gamma$ & 0.99 \\
Expectile $\tau$ & 0.9 \\
Target networks & EMA, rate 0.995 \\
AWR $\beta$ & 3 (GCIVL), 10 (GCIQL); weights clipped at 100 \\
Value goals & random 0.3, geometric future 0.5, current 0.2 \\
Policy goals & GCBC: uniform future 1.0; GCIVL, GCIQL: random 0.5, uniform future 0.5 \\
Geometric future goals & discount 0.99 \\
\bottomrule
\end{tabularx}
\end{table}

GCBC, GCIVL and GCIQL learn a goal-conditioned policy
$\pi(\boldsymbol{a} \mid \boldsymbol{s}, \boldsymbol{g})$ from the training episodes of LeWM.
They use no world model at test time, so they test whether a policy learned directly from
the data does better than planning. We use the objectives and hyperparameters of the
stable-worldmodel implementations~\cite{maes2026swm}, but train MLPs on the frozen LeWM
embeddings instead of images (\cref{tab:supp-gcrl}). The state $\boldsymbol{s}_t$ stacks the
embeddings of the last three frames, the goal $\boldsymbol{g}$ is the embedding of goal frame,
and $\boldsymbol{a}_t$ is the logged action of step $t$. Goals are relabeled from the data as
in OGBench~\cite{park2025ogbench}. The reward $r(\boldsymbol{s}, \boldsymbol{g})$ is $0$ when the current frame equals the goal frame
and $-1$ otherwise, and the value target stops bootstrapping at the goal.

\paragraph{GCBC~\cite{ghosh2021gcsl}.} GCBC imitates the actions of the data, with a later
frame of the same episode as the goal,
\begin{equation}
  \mathcal{L}_{\text{GCBC}}(\pi) = -\mathbb{E}_{(\boldsymbol{s}, \boldsymbol{a}, \boldsymbol{g}) \sim \mathcal{D}}
  \big[\log \pi(\boldsymbol{a} \mid \boldsymbol{s}, \boldsymbol{g})\big].
  \label{eq:supp-gcbc}
\end{equation}

\paragraph{GCIVL~\cite{park2023hiql}.} GCIVL learns a goal-conditioned state value by
expectile regression~\cite{kostrikov2022iql} and extracts the policy with advantage-weighted
regression (AWR)~\cite{peng2019awr},
\begin{equation}
  \begin{aligned}
    \mathcal{L}(V) &= \mathbb{E}\big[\ell^2_{\tau}\big(r(\boldsymbol{s}, \boldsymbol{g})
      + \gamma \bar{V}(\boldsymbol{s}', \boldsymbol{g}) - V(\boldsymbol{s}, \boldsymbol{g})\big)\big],\\
    \mathcal{L}(\pi) &= -\mathbb{E}\big[\min\big(e^{\beta A},\, 100\big)
      \log \pi(\boldsymbol{a} \mid \boldsymbol{s}, \boldsymbol{g})\big],
  \end{aligned}
  \label{eq:supp-gcivl}
\end{equation}
where $\ell^2_{\tau}(u) = |\tau - \mathbb{1}(u < 0)|\, u^2$ is the expectile loss, $\bar{V}$
is a target network, and $A = V(\boldsymbol{s}', \boldsymbol{g}) - V(\boldsymbol{s}, \boldsymbol{g})$.

\paragraph{GCIQL~\cite{kostrikov2022iql}.} GCIQL learns an action value $Q$ alongside $V$,
\begin{equation}
  \begin{aligned}
    \mathcal{L}(V) &= \mathbb{E}\big[\ell^2_{\tau}\big(\bar{Q}(\boldsymbol{s}, \boldsymbol{a}, \boldsymbol{g})
      - V(\boldsymbol{s}, \boldsymbol{g})\big)\big],\\
    \mathcal{L}(Q) &= \mathbb{E}\big[\big(r(\boldsymbol{s}, \boldsymbol{g})
      + \gamma \bar{V}(\boldsymbol{s}', \boldsymbol{g}) - Q(\boldsymbol{s}, \boldsymbol{a}, \boldsymbol{g})\big)^2\big],
  \end{aligned}
  \label{eq:supp-gciql}
\end{equation}
and extracts the policy as GCIVL does, with a larger $\beta$.

\FloatBarrier
\section{Branching with the True Simulator}
\label{app:exact}
Branching pays only if the agent observes something before an irreversible outcome that changes its best continuation (\cref{prop:info}). With a learned model, a planner can miss such an observation because the predictions are wrong. To test branching without model error, we replace LeWM with the simulator and the learned time-to-goal with the true time to the goal, on TwoRoom-Trap, TwoRoom-LockedDoor and TwoRoom-Slip.

\paragraph{Setup.} Each of the $K = 64$ futures is a rollout of the simulator with its own noise. On TwoRoom-LockedDoor, each future also draws the locked door until the lamps show it. A future costs its steps to the goal, plus the steps it still needs at full speed after the horizon. A fall costs more than any future that does not fall, and each step past the budget costs 5 more. CEM samples 300 candidates and refits to the 30 best for 30 iterations, and the agent replans after every chunk. The horizon is 5 chunks (10 on TwoRoom-LockedDoor), and the budget is 50 steps (25 on TwoRoom-Slip). Branching follows CoLaP, but groups the futures by what the agent observes. For comparison, we also score futures by the sum of squared distances to the goal, with the mean or CVaR$_{0.25}$. Each cell runs 50 episodes on four seeds.

\begin{table}[!htbp]
\centering
\caption{\textbf{Branching with the true simulator.} We report success rate (\%) over four seeds, with 50 episodes per seed. In brackets are the episodes (\%) that fell on TwoRoom-Trap and that visited the switch on TwoRoom-LockedDoor.}
\label{tab:supp-exact}
\footnotesize\ttfamily
\setlength{\tabcolsep}{4pt}
\resizebox{\linewidth}{!}{%
\begin{tabular}{llccc}
\toprule
Cost & Planner & Trap & LockedDoor & Slip \\
\midrule
\multirow{2}{*}{Expected time} & Open loop & 100.0 (0.0) & 59.0 (0.0)  & 100.0 \\
                               & Branching & 100.0 (0.0) & 87.0 (73.5) & 100.0 \\
\midrule
\multirow{3}{*}{Sum of distances} & Open loop       & 82.0 (18.0) & 36.0 (0.0)  & 99.5 \\
                                  & Branching, mean & 83.0 (17.0) & 87.5 (75.0) & 99.5 \\
                                  & Branching, CVaR & 96.0 (4.0)  & 55.0 (55.0) & 99.5 \\
\bottomrule
\end{tabular}}
\end{table}

\paragraph{Results.} On TwoRoom-LockedDoor, branching raises success from 59.0 to 87.0\% (\cref{tab:supp-exact}), with 59 episodes won and 3 lost. It visits the switch in 73.5\% of the episodes, while open loop never does. The switch shows the lock before the agent tries a door, so it is the observation that \cref{prop:info} asks for (\cref{main-fig:components}c). On TwoRoom-Trap, a fall comes without warning, and both planners reach 100\%. So, branching neither helps nor hurts there. With the sum of distances, no planner works on both tasks. Branching with the mean reaches 87.5\% on TwoRoom-LockedDoor but falls in 17.0\% of the TwoRoom-Trap episodes, and CVaR falls in only 4.0\% but reaches 55.0\% on TwoRoom-LockedDoor. The expected time handles both tasks with one objective.

\FloatBarrier
\section{Additional Experiments}
\label{app:results}
We run further experiments on our model. As in \cref{main-tab:main}, every success rate is average over four seeds.
Each seed contains 50 episodes. Every ablation changes one part of CoLaP and keeps the rest. 
Unless stated otherwise, all tasks use $K = 8$ futures.

\subsection{Clipped actions}
\label{app:abl-clip}
CoLaP does not clip the CEM candidates to the action range (\cref{app:impl-planner}). \Cref{tab:supp-clip} clips each candidate, so that no planned or executed action leaves the range. Clipping changes CoLaP by $-2.5$ to $-0.5$ points. On average, clipping hurts one predicted future and MC-dropout significantly. So, CoLaP keeps its lead even when the actions are clipped.

\begin{table}[!htbp]
\centering
\caption{\textbf{Clipped actions.} We report success rate (\%) and standard deviation over four seeds, without and with every CEM candidate clipped to the action range.}
\label{tab:supp-clip}
\footnotesize\ttfamily
\setlength{\tabcolsep}{3pt}
\resizebox{\linewidth}{!}{%
\begin{tabular}{lcccccc}
\toprule
 & \multicolumn{2}{c}{One future} & \multicolumn{2}{c}{MC-dropout} & \multicolumn{2}{c}{CoLaP} \\
\cmidrule(lr){2-3}\cmidrule(lr){4-5}\cmidrule(lr){6-7}
Task & No clip & Clip & No clip & Clip & No clip & Clip \\
\midrule
TwoRoom-Trap   & \scell{77.5}{5.7} & \scell{60.0}{5.9} & \scell{96.5}{1.9} & \scell{74.0}{6.5} & \sbest{99.5}{1.0} & \scell{97.5}{1.9} \\
TwoRoom-Bridge & \scell{57.5}{2.5} & \scell{54.5}{5.0} & \scell{59.5}{4.4} & \scell{55.5}{4.4} & \sbest{83.5}{1.9} & \scell{81.0}{6.2} \\
Reacher-Thin   & \scell{67.5}{2.5} & \scell{69.0}{2.0} & \scell{71.5}{1.9} & \scell{69.0}{2.6} & \sbest{92.0}{5.9} & \scell{91.5}{7.2} \\
Reacher-Hazard & \scell{69.0}{6.2} & \scell{69.5}{6.0} & \scell{73.0}{7.4} & \scell{71.0}{4.2} & \sbest{90.5}{7.2} & \scell{88.5}{6.2} \\
\cmidrule(lr){1-7}
Average        & 67.9 & 63.2 & 75.1 & 67.4 & \sbestavg{91.4} & 89.6 \\
\bottomrule
\end{tabular}}
\end{table}

\subsection{Equal planning time}
\label{app:abl-compute}
CoLaP scores 8 futures of two plans, so one re-plan takes longer than for one predicted future. To test whether its gain comes from this extra compute, we give one predicted future $N$ CEM samples instead of 300, so that one re-plan takes as long as one re-plan of CoLaP with $K = 8$ (\cref{main-fig:design}c). It keeps $N/10$ elites, as CoLaP keeps 30 of 300. We set $N$ by timing the first three re-plans of 50 episodes on an idle GPU, and interpolate between a run with 3{,}000 samples and a check run near the target. This gives $N = 6{,}000$ to $6{,}600$ (\cref{tab:supp-compute}).

\begin{table}[!htbp]
\centering
\caption{\textbf{Equal planning time.} We report success rate (\%) and standard deviation over four seeds for one predicted future with 300 and $N$ CEM samples. CoLaP is the column of \cref{main-tab:main}, as in \cref{main-fig:design}c. The last column gives the falls (\%) of one predicted future with 300 and $N$ samples.}
\label{tab:supp-compute}
\footnotesize\ttfamily
\setlength{\tabcolsep}{3pt}
\resizebox{\linewidth}{!}{%
\begin{tabular}{lccccc}
\toprule
 & & \multicolumn{2}{c}{One future} & & Falls \\
\cmidrule(lr){3-4}
Task & $N$ & 300 & $N$ & CoLaP & 300 / $N$ \\
\midrule
TwoRoom-Trap   & 6{,}600 & \scell{77.5}{5.7} & \scell{87.5}{3.0} & \sbest{99.5}{1.0} & 22.5 / 12.5 \\
TwoRoom-Bridge & 6{,}500 & \scell{57.5}{2.5} & \scell{59.0}{6.2} & \sbest{83.5}{1.9} & 42.5 / 41.0 \\
Reacher-Thin   & 6{,}600 & \scell{67.5}{2.5} & \scell{66.0}{3.3} & \sbest{92.0}{5.9} & 32.5 / 30.5 \\
PushT-Pusher   & 6{,}000 & \scell{36.0}{2.8} & \scell{34.5}{3.8} & \sbest{50.5}{5.4} & 24.0 / 25.0 \\
\cmidrule(lr){1-6}
Average        &         & 59.6 & 61.8 & \sbestavg{81.4} & \\
\bottomrule
\end{tabular}}
\end{table}

The larger search raises one predicted future from 59.6 to 61.8\% on average, while CoLaP reaches 81.4\% on the given tasks. It helps only on TwoRoom-Trap, by 10.0 points ($p = 4 \times 10^{-5}$), and changes the other tasks by $-1.5$ to $+1.5$ points. Its falls barely change on TwoRoom-Bridge (42.5 to 41.0\%) and Reacher-Thin (32.5 to 30.5\%). More search finds a better plan for the single prediction, but that prediction does not show the irreversible outcome. So, CoLaP gains from what it scores, not from more compute. This matches \cref{thm:meantrap}, whose bound holds for any search budget.


\subsection{The \texorpdfstring{$K = 32$}{K = 32} versions}
\label{app:abl-k32}

CoLaP and the ablations of the main paper use $K = 8$ futures. \Cref{tab:supp-k32} repeats \cref{main-tab:components} and \cref{main-fig:design}b with $K = 32$, and the conclusions hold. Without the time-to-goal, success drops by 12.5 to 38.5 points, and without the sampler by 10.5 to 26.0 points. Our time-to-goal is the best cost on three of the four tasks, and the symmetric cost leads on TwoRoom-Bridge (78.0 against 76.0\%). Branching adds 0.5 to 17.0 points, more than with $K = 8$.

\begin{table*}[!htbp]
\centering
\caption{\textbf{Ablations with $K = 32$.} We report success rate (\%) and standard deviation over four seeds. The first four columns are CoLaP and CoLaP without one component, as in \cref{main-tab:components}. The last three columns replace our time-to-goal in CoLaP without branching, as in \cref{main-fig:design}b. Latent is the latent distance of LeWM.}
\label{tab:supp-k32}
\footnotesize\ttfamily
\setlength{\tabcolsep}{0pt}
\begin{tabular*}{\textwidth}{@{\extracolsep{\fill}}lccccccc@{}}
\toprule
 & & \multicolumn{3}{c}{CoLaP without} & \multicolumn{3}{c}{No branching, other cost} \\
\cmidrule(lr){3-5}\cmidrule(lr){6-8}
Task & CoLaP & Sampler & Time-to-goal & Branching & Latent & Symmetric & TEMPO \\
\midrule
TwoRoom-Trap   & \sbest{99.0}{1.2} & \scell{73.0}{7.0} & \scell{85.5}{2.5} & \scell{98.5}{1.0} & \scell{88.5}{3.4} & \scell{52.5}{5.5} & \scell{64.0}{11.2} \\
TwoRoom-Bridge & \sbest{80.5}{6.6} & \scell{63.0}{7.4} & \scell{68.0}{2.8} & \scell{76.0}{7.1} & \scell{56.5}{3.8} & \scell{78.0}{5.7} & \scell{75.5}{2.5} \\
Reacher-Thin   & \sbest{88.5}{4.1} & \scell{71.5}{3.4} & \scell{50.0}{4.3} & \scell{79.0}{5.3} & \scell{49.5}{4.1} & \scell{56.5}{3.4} & \scell{45.0}{1.2} \\
Reacher-Hazard & \sbest{83.5}{9.1} & \scell{73.0}{7.6} & \scell{49.0}{2.6} & \scell{66.5}{4.1} & \scell{44.0}{2.3} & \scell{50.0}{1.6} & \scell{37.5}{3.0} \\
\cmidrule(lr){1-8}
Average        & \sbestavg{87.9} & 70.1 & 63.1 & 80.0 & 59.6 & 59.2 & 55.5 \\
\bottomrule
\end{tabular*}
\end{table*}

\subsection{Second World Model: PLDM}
\label{app:pldm}
CoLaP never changes the world model. To test whether it carries over to another one, we replace LeWM with PLDM~\cite{sobal2025pldm} and run CoLaP on all nine tasks without any change. We use the PLDM of stable-worldmodel~\cite{maes2026swm}. It has the same encoder, predictor and embedding size as our LeWM, but replaces SIGReg with the variance and covariance terms of VICReg~\cite{bardes2022vicreg} and two temporal terms. We train it on the same training episodes of LeWM, train the sampler and the time-to-goal on its embeddings with the same method, and plan with the same settings. Nothing is tuned for PLDM.

\begin{table}[!htbp]
\centering
\caption{\textbf{CoLaP on two world models.} We report success rate (\%) and standard deviation over four seeds, with 50 episodes per seed. CoLaP uses $K = 8$ futures on both.}
\label{tab:supp-pldm}
\footnotesize\ttfamily
\setlength{\tabcolsep}{0pt}
\begin{tabular*}{\linewidth}{@{\extracolsep{\fill}}lcc@{}}
\toprule
Task & PLDM & LeWM \\
\midrule
TwoRoom-Trap       & \scell{84.5}{7.0}  & \scell{99.5}{1.0} \\
TwoRoom-Bridge     & \scell{87.5}{4.4}  & \scell{83.5}{1.9} \\
TwoRoom-Slip       & \scell{99.0}{1.2}  & \scell{100.0}{0.0} \\
TwoRoom-LockedDoor & \scell{56.0}{12.3} & \scell{71.0}{6.1} \\
Reacher-Thin       & \scell{94.0}{6.3}  & \scell{92.0}{5.9} \\
Reacher-Hazard     & \scell{89.0}{4.8}  & \scell{90.5}{7.2} \\
PushT-Pit          & \scell{12.5}{1.0}  & \scell{13.5}{4.4} \\
PushT-Pusher       & \scell{46.5}{4.4}  & \scell{50.5}{5.4} \\
Cube-Drop          & \scell{0.5}{1.0}   & \scell{0.5}{1.0} \\
\cmidrule(lr){1-3}
Average            & 63.3               & 66.8 \\
\bottomrule
\end{tabular*}
\end{table}

\Cref{tab:supp-pldm} shows that CoLaP performs about the same on both world models. Its average is 63.3\% on PLDM and 66.8\% on LeWM, and on seven of the nine tasks the two differ by at most 4.0 points. The largest gaps are on TwoRoom-Trap and TwoRoom-LockedDoor, where PLDM is 15.0 points lower. Therefore, CoLaP carries over to a different world model without any changes.

\FloatBarrier
\section{Additional Analyses}
\label{app:analyses}

\subsection{Do the sampled futures show the fall?}
\label{app:ana-calib}
CoLaP assumes that the single prediction of LeWM hides the irreversible outcome, while sampled futures show it (\cref{thm:meantrap}). We test this on risky windows of held-out episodes, where the agent or the object it moves is near the hazard and has not fallen yet. A linear probe labels each predicted next embedding as fallen or not, and the fall share $q$ is the fraction of predictions labeled fallen. We compare one predicted future, 32 MC-dropout passes and 64 draws of our sampler (\cref{tab:supp-calib}).

\begin{table}[!htbp]
\centering
\caption{\textbf{Do the sampled futures show the fall?} Windows whose predictions hold both outcomes, $0 < q < 1$ (Both, \%), and the Brier score and AUROC of the fall share $q$ against the real outcome. Under each task are its risky windows and their real fall rate.}
\label{tab:supp-calib}
\footnotesize\ttfamily
\setlength{\tabcolsep}{0pt}
\begin{tabular*}{\linewidth}{@{\extracolsep{\fill}}llccc@{}}
\toprule
Task & Futures & Both & Brier $\downarrow$ & AUROC $\uparrow$ \\
\midrule
\multirow{3}{*}{\shortstack[l]{TwoRoom-Trap\\4{,}222, 7.2\%}}
 & One future  & 0.0  & 0.092 & 0.64 \\
 & MC-dropout  & 11.9 & 0.075 & 0.77 \\
 & Our sampler & 31.0 & \sbestavg{0.057} & \sbestavg{0.88} \\
\midrule
\multirow{3}{*}{\shortstack[l]{TwoRoom-Bridge\\9{,}727, 5.5\%}}
 & One future  & 0.0  & 0.068 & 0.59 \\
 & MC-dropout  & 8.7  & 0.055 & 0.71 \\
 & Our sampler & 31.0 & \sbestavg{0.043} & \sbestavg{0.91} \\
\midrule
\multirow{3}{*}{\shortstack[l]{Reacher-Hazard\\19{,}389, 2.3\%}}
 & One future  & 0.0  & 0.038 & 0.71 \\
 & MC-dropout  & 2.4  & 0.034 & 0.77 \\
 & Our sampler & 10.4 & \sbestavg{0.019} & \sbestavg{0.94} \\
\midrule
\multirow{3}{*}{\shortstack[l]{PushT-Pusher\\32{,}413, 1.5\%}}
 & One future  & 0.0  & 0.018 & 0.63 \\
 & MC-dropout  & 1.3  & 0.017 & 0.69 \\
 & Our sampler & 5.9  & \sbestavg{0.014} & \sbestavg{0.85} \\
\bottomrule
\end{tabular*}
\end{table}

\begin{figure*}[!htbp]
\centering
\includegraphics[width=\textwidth]{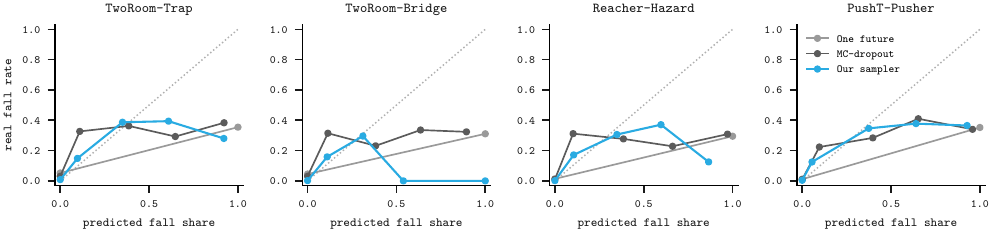}
\caption{\textbf{Calibration of the fall share.} The real fall rate against the fall share $q$ on the risky windows, in one bin for $q = 0$ and four bins of width 0.25. The dotted line is perfect calibration.}
\label{fig:supp-calib}
\end{figure*}

Our sampler has the lowest Brier score and the highest AUROC on all four tasks, and its fall share is calibrated (\cref{fig:supp-calib}). One predicted future never holds both outcomes, and dropout rarely does. On the windows that fell, the probe labels the single prediction as not fallen (logit $-1.4$ to $-1.7$, against 4.2 to 5.6 for the real frame). This is the mean trap of \cref{thm:meantrap}, measured.

\subsection{Test-time hazard probability}
\label{app:ana-hazardp}
LeWM, the sampler, and the time-to-goal are trained on data where one crossing fails with probability 0.5 (\cref{app:setup-envs}). \Cref{tab:supp-hp} tests them at 0.25 and 0.75 without retraining. CoLaP stays ahead of one predicted future at every probability, and its gain grows with it, from 7.5 to 23.0 points on TwoRoom-Trap and from 11.5 to 37.5 points on Reacher-Thin.

\begin{table}[!htbp]
\centering
\caption{\textbf{Test-time hazard probability.} We report success rate (\%) and standard deviation over four seeds when one crossing fails with probability $p$ at test time. All components are trained at $p = 0.5$.}
\label{tab:supp-hp}
\footnotesize\ttfamily
\setlength{\tabcolsep}{0pt}
\begin{tabular*}{\linewidth}{@{\extracolsep{\fill}}lcccc@{}}
\toprule
Task & $p$ & One future & CoLaP & Gain \\
\midrule
\multirow{3}{*}{TwoRoom-Trap}
 & 0.25 & \scell{92.0}{1.6} & \sbest{99.5}{1.0} & +7.5 \\
 & 0.50 & \scell{77.5}{5.7} & \sbest{99.5}{1.0} & +22.0 \\
 & 0.75 & \scell{70.5}{7.5} & \sbest{93.5}{1.9} & +23.0 \\
\midrule
\multirow{3}{*}{Reacher-Thin}
 & 0.25 & \scell{86.5}{2.5} & \sbest{98.0}{4.0} & +11.5 \\
 & 0.50 & \scell{67.5}{2.5} & \sbest{92.0}{5.9} & +24.5 \\
 & 0.75 & \scell{50.0}{6.3} & \sbest{87.5}{5.3} & +37.5 \\
\bottomrule
\end{tabular*}
\end{table}


\subsection{When does CoLaP pay?}
\label{app:ana-when}
\Cref{tab:supp-when} tests whether CoLaP gains where the hazard hurts. The loss is what the hazard costs one predicted future, its success with the hazard switched off minus its success in \cref{main-tab:main}. The gain is CoLaP minus one predicted future in \cref{main-tab:main}. The gain tracks the loss (Spearman $\rho = 0.93$), and where the loss exceeds 5 points, CoLaP wins back 61 to 100\% of it. On PushT-Pit and Cube-Drop, one predicted future fails even without the hazard, so there is nothing to win back. 
So, CoLaP fixes the irreversible outcome, not other planning failures.

\begin{table}[!htbp]
\centering
\caption{\textbf{When does CoLaP pay?} Success points (\%) on the seven hazard tasks. Off is one predicted future with the hazard switched off. Loss is Off minus one predicted future in \cref{main-tab:main}, gain is CoLaP minus one predicted future in \cref{main-tab:main}, and won back is the gain over the loss where the loss exceeds 5 points.}
\label{tab:supp-when}
\footnotesize\ttfamily
\setlength{\tabcolsep}{0pt}
\begin{tabular*}{\linewidth}{@{\extracolsep{\fill}}lcccc@{}}
\toprule
Task & Off & Loss & Gain & Won back \\
\midrule
TwoRoom-Trap   & 100.0 & 22.5 & 22.0 & 98  \\
TwoRoom-Bridge & 100.0 & 42.5 & 26.0 & 61  \\
Reacher-Thin   & 100.0 & 32.5 & 24.5 & 75  \\
Reacher-Hazard & 99.5  & 30.5 & 21.5 & 70  \\
PushT-Pit      & 19.5  & 5.0  & -1.0 & --  \\
PushT-Pusher   & 50.5  & 14.5 & 14.5 & 100 \\
Cube-Drop      & 0.0   & 0.0  & 0.5  & --  \\
\bottomrule
\end{tabular*}
\end{table}


\subsection{Qualitative example}
\label{app:ana-qual}
\begin{figure*}[!htbp]
\centering
\includegraphics[width=\textwidth]{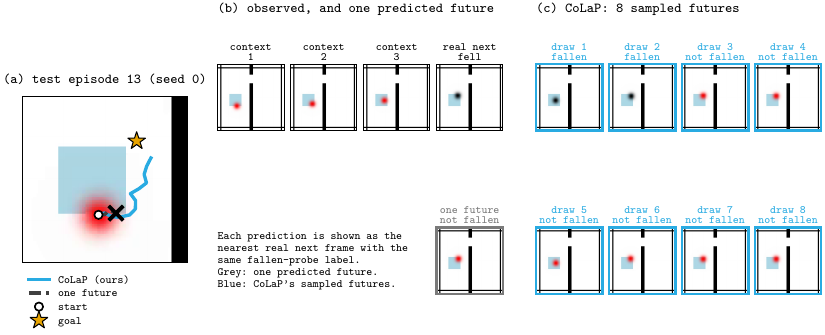}
\caption{\textbf{Qualitative example on TwoRoom-Trap.} (a) Test episode 13 of seed 0, the first episode where one predicted future falls and CoLaP succeeds, cropped around the ice. (b) A held-out moment on the ice where the agent fell. We show the three context frames, the real next frame and the prediction of LeWM. (c) The 8 samples of CoLaP for the same moment. The probe of \cref{app:ana-calib} labels each prediction, and we show the nearest real frame with that label.}
\label{fig:supp-qual}
\end{figure*}

\Cref{fig:supp-qual} shows the mean trap on TwoRoom-Trap. In a test episode, one predicted future walks onto the ice and falls, while CoLaP goes around it. 
We also show a held-out moment on the ice where the agent fell. The single prediction of LeWM shows no fall, while 2 of the 8 samples of CoLaP show a fall. This share of 0.25 is close to the real fall rate on the ice, 0.29.

\FloatBarrier

\end{document}